\documentclass[lettersize,journal]{IEEEtran}

\usepackage{hyperref}
\usepackage[dvipsnames]{xcolor}

\usepackage{amsmath,amsfonts}
\usepackage{algorithmic}
\usepackage{algorithm}
\usepackage{array}
\usepackage[caption=false,font=footnotesize,labelfont=rm,textfont=rm]{subfig}
\usepackage{textcomp}
\usepackage{stfloats}
\usepackage{url}
\usepackage{verbatim}
\usepackage{graphicx}
\usepackage{cite}
\usepackage{cleveref}
\usepackage{booktabs}

\newtheorem{theorem}{Theorem}
\newtheorem{proof}{Proof}[section]
\newtheorem{definition}{Definition}
\newtheorem{lemma}{Lemma}
\usepackage{multirow}
\usepackage{multicol}
\usepackage{color}

\begin{document}

\title{Diverse Target and Contribution Scheduling for Domain Generalization}

\author{
Shaocong Long$^\star$, 
        Qianyu Zhou$^\star$,
        Chenhao Ying, 
        Lizhuang Ma,
        Yuan Luo$^\dag$
\thanks{Shaocong Long, Qianyu Zhou, Chenhao Ying, Lizhuang Ma, Yuan Luo are with the Department of Computer Science and Engineering, Shanghai Jiao Tong University, Shanghai 200240, China~(email: \{longshaocong, zhouqianyu, yingchenhao, yuanluo\}@sjtu.edu.cn and ma-lz@cs.sjtu.edu.cn).}
\thanks{$^\star$Equal contribution, $^\dag$Corresponding author.}
}


\IEEEpubid{0000--0000/00\$00.00~\copyright~2023 IEEE}
\IEEEpubidadjcol

\maketitle

\begin{abstract}
Generalization under the distribution shift has been a great challenge in computer vision. The prevailing practice of directly employing the one-hot labels as the training targets in domain generalization~(DG) can lead to gradient conflicts, making it insufficient for capturing the intrinsic class characteristics and hard to increase the intra-class variation. Besides, existing methods in DG mostly overlook the distinct contributions of source (seen) domains, resulting in uneven learning from these domains. To address these issues, we firstly present a theoretical and empirical analysis on the existence of gradient conflicts in DG, unveiling the previously unexplored relationship between distribution shifts and gradient conflicts during the optimization process. In this paper, we present a novel perspective of DG from the empirical source domain's risk and propose a new paradigm for DG called Diverse Target and Contribution 
Scheduling~(DTCS). DTCS comprises two innovative modules: Diverse Target Supervision~(DTS) and Diverse Contribution Balance~(DCB), with the aim of addressing the limitations associated with the common utilization of one-hot labels and equal contributions for source domains in DG. In specific, DTS employs distinct soft labels as training targets to account for various feature distributions across domains and thereby mitigates the gradient conflicts, and DCB dynamically balances the contributions of source domains by ensuring a fair decline in losses of different source domains. Extensive experiments with analysis on four benchmark datasets show that the proposed method achieves a competitive performance in comparison with the state-of-the-art approaches, demonstrating the effectiveness and advantages of the proposed DTCS.
\end{abstract}


\section{Introduction}

Machine learning has achieved remarkable success across a wide range of areas, including image classification~\cite{krizhevsky2012imagenet, he2016deep, dosovitskiy2020image, deng2022dynamic,feng2022dmt,song2023rethinking}, object detection~\cite{ren2015faster, lin2017feature, redmon2016you, lim2023ernet, yang2022efficient,he2021end,zhou2022transvod,xu2021semi}, and image segmentation~\cite{long2015fully, ronneberger2015u, chen2018encoder, ramesh2022hierarchical, yuan2022birds}. However, the real-world applicability of machine learning models faces significant challenges when the assumption of independent and identically distributed (i.i.d.) does not hold~\cite{beery2018recognition,li2020domain,ding2022word, deng2022dynamic, zhao2022style}. The dilemma mainly arises from the inherent vulnerability  of neural networks to spurious relationships, leading them to rely on superficial shortcuts instead of capturing genuine underlying patterns~\cite{geirhos2020shortcut,zhang2021can}. For instance, a model may associate images with pastures with cows, disregarding the actual animals~\cite{beery2018recognition}, thereby suffering heavily in generalizing out of samples.

Numerous endeavors have been dedicated to mitigating the risks caused by distribution shifts. With access to unlabeled data from target domains, unsupervised domain adaptation~(UDA)~\cite{kong2022partial,stojanov2021domain, yuan2022birds, deng2022dynamic, wang2022cluster,gu2021pit,zhou2022context,zhou2022generative,zhou2022uncertainty,zhou2022domain,zhou2023self} aims to adapt a model trained on labeled source domains to accommodate unlabeled target domains. However, the trained models still require adaptations when novel environments occur, which is time-consuming. In this study, we focus on the task of domain generalization~(DG)~\cite{gulrajani2020search,wang2022generalizing, wang2022contrastive, xia2023generative,zhou2023instance,zhou2022adaptive} which is a more challenging task to diminish the influence of distribution shift without the luxury of accessing target domain.

\IEEEpubidadjcol

Many existing methods~\cite{ganin2016domain,li2018deep,zhao2020domain,yao2022pcl} in DG attempt to acquire a domain-invariant representation capable of generalizing to novel domains by suppressing the domain-specific features such as styles, visual angles, and others. However, due to the existence of hard samples, the pursuit of invariant features and indiscriminate elimination of domain-specific information can lead to adverse effects on within-class variations and amplify the empirical source risk, ultimately undermining the model's generalization performance~\cite{mahajan2021domain, yao2022pcl}. 

\begin{figure}[t!]
    \centering
    
    \subfloat[Same targets]{\includegraphics[width=1.8in]{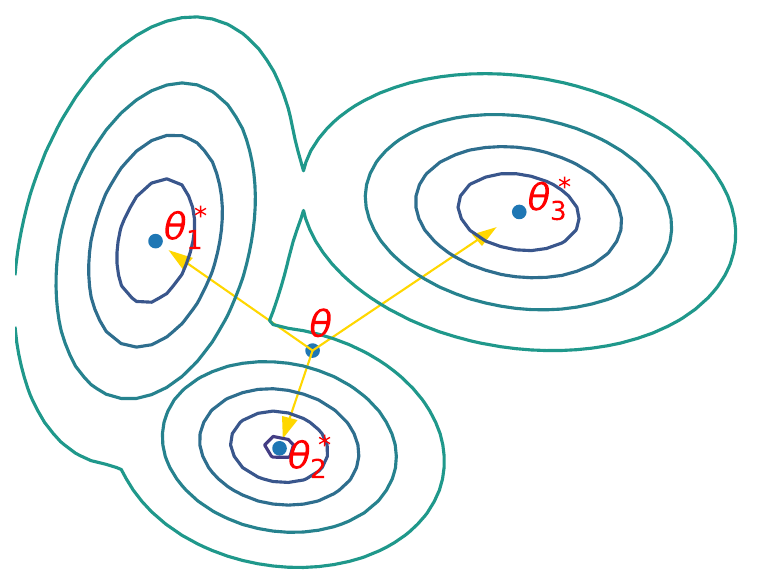}}
        \subfloat[Diverse targets]{\includegraphics[width=1.8in]{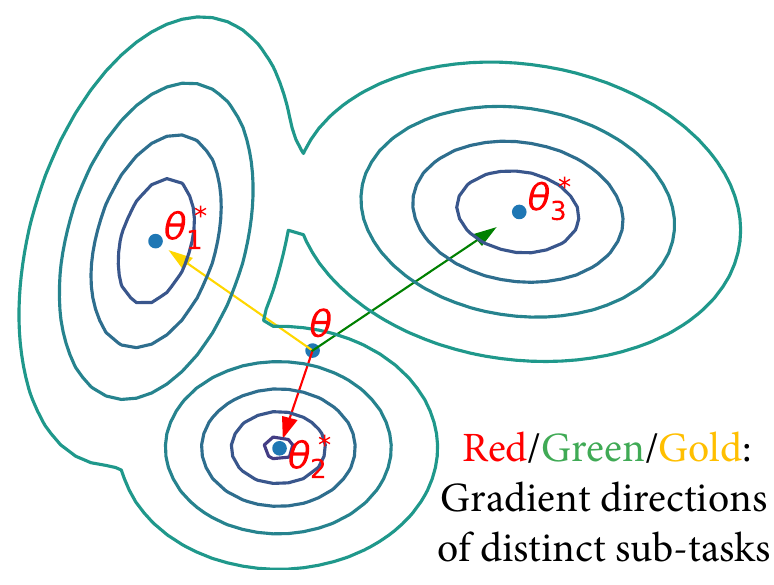}}
    
    \caption{\textbf{The origin of gradient conflicts in DG and our mitigation strategy.} $\theta_1^*$, $\theta_2^*$, $\theta_3^*$ denote the optimal solutions in different source domains, and $\theta$ represents the current model parameter. (a) In DG, the probability density functions exhibit variations across domains, resulting in a multi-modal distribution for one class and thus causing disparate gradient descent directions across domains. The gradients in different domains may contradict with each other, hindering the optimization process. (b) Introducing diverse targets for source domains experiencing distribution shifts, our proposed strategy aims to implicitly divide the overall optimization process into distinct sub-tasks that do not interface each other.}
    \label{contour}
\end{figure}

\begin{figure}[t]
  \centering
  \includegraphics[width=0.9\linewidth]{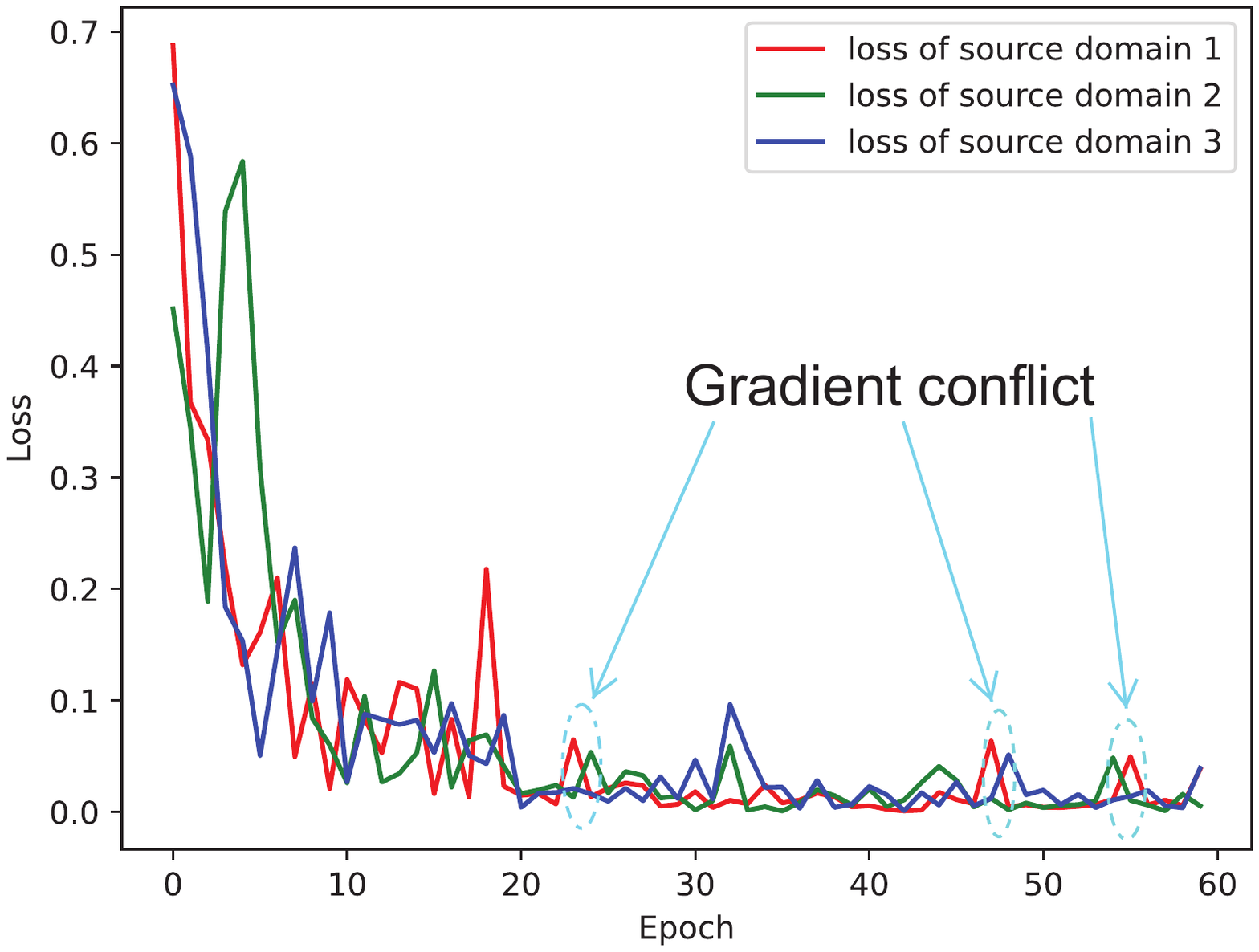}
  \setlength{\abovecaptionskip}{-0.05cm}
  \caption{The training losses of different source domains for ERM with the pre-trained ResNet-18 as the backbone and `Photo' as the target domain, source domain 1, 2, and 3 denote `Art painting', `Cartoon', and `Sketch', respectively.}
  \label{ERM_loss}
\end{figure}

\begin{figure}[t]
  \centering
   \includegraphics[width=0.9\linewidth]{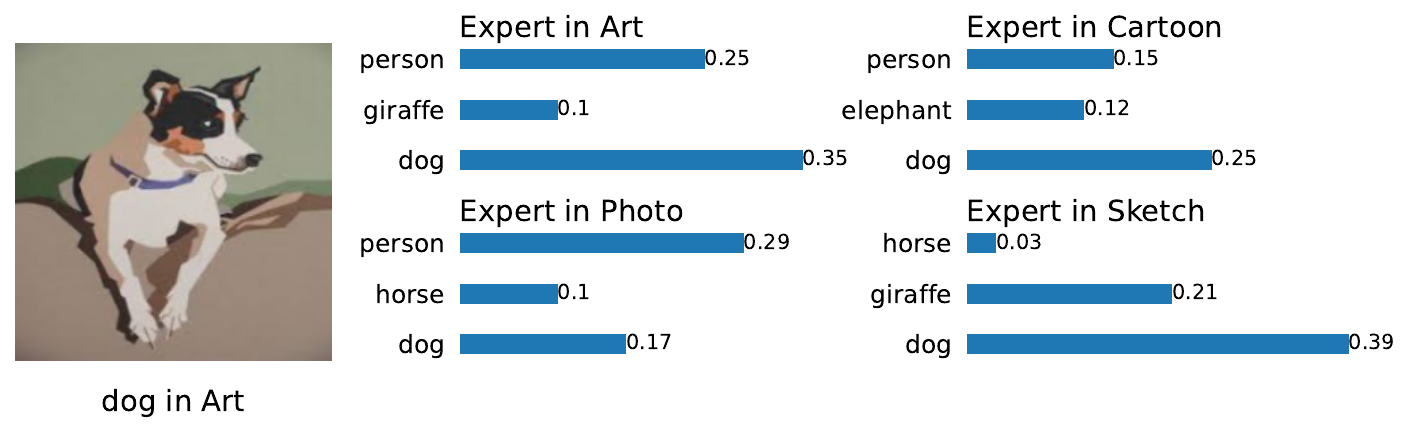}
   \caption{The softmax logits (the top-3 predicted classes and their corresponding confidences) generated by the specialist models trained in `Art', `Cartoon', `Photo', and `Sketch', respectively. The dog's picture is sampled from `Art'.}
   \label{logits}
\end{figure}

\begin{figure}[t]
    \centering
    \includegraphics[width=0.99\linewidth]{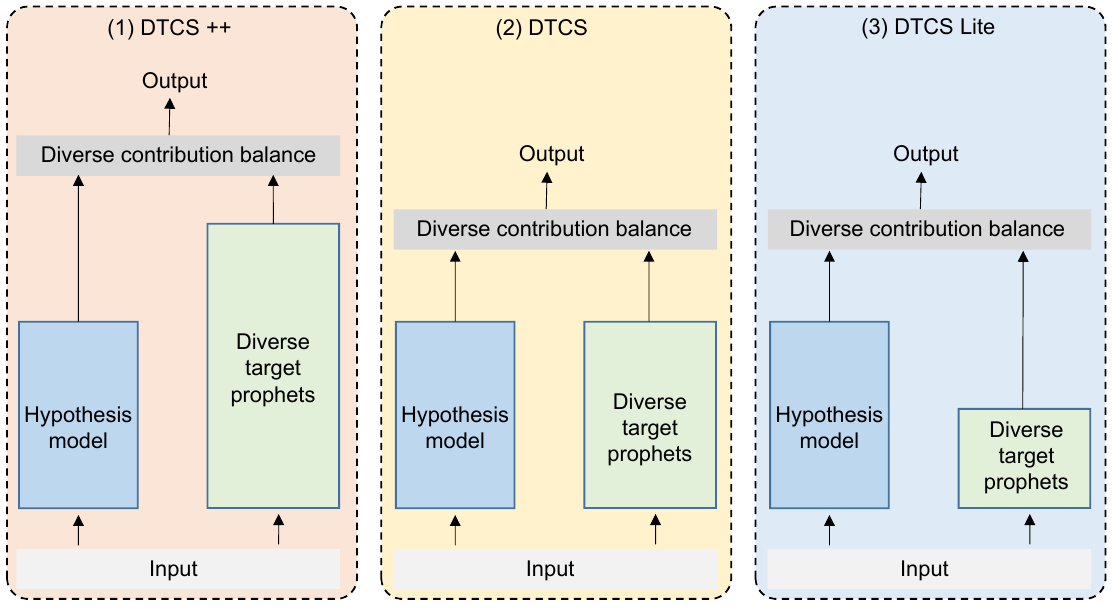}
    \caption{Illustration of the proposed Diverse Target and Contribution Scheduling~(DTCS). We propose three distinct paradigms for DTCS, with diverse target prophets ranging from complex to simple, while ensuring negligible performance degradation.}
    \label{archi}
\end{figure}

The motivation for domain-invariant learning in DG arises from the requirement to align samples undergoing distribution shifts across domains to conform to the same one-hot labels. However, the mainstream methods~\cite{vapnik1999overview, ganin2016domain,li2018deep,zhao2020domain,yao2022pcl, li2018domain, yan2020improve, wang2022contrastive, meng2022attention, li2022sparse, li2022invariant} in DG tend to overlook a crucial aspect: relying solely on one-hot labels may lose sight of underlying data semantics and thus is insufficient to capture the diverse class characteristics, making it tough for the model to fit examples with various distributions to the same one-hot label. In specific, two key limitations arise from using one-hot labels as training targets for samples across source domains: (\romannumeral1) Employing identical one-hot labels for samples from diverse source domains can impede the training process due to conflicts in the gradient descent directions caused by distribution shifts. Just as shown in Fig.~\ref{contour}, the multi-modal distribution for one class results in distinct sub-optimal solutions for different source domains, leading to disparate gradient descent directions. Such conflicting directions hamper optimization and impede convergence to global optimum, a common occurrence in real-world scenarios. Depicting the losses of different source domains respectively, Fig.~\ref{ERM_loss} is consistent with the claim, and the wild fluctuations in the losses indicate an unstable training process. Furthermore, as depicted by the dashed ellipse, the declining loss in one domain may result in an increasing loss in another domain, highlighting the contradictory gradient descent directions across domains during training. (\romannumeral2) Employing one-hot labels as training targets falls short in diversifying the variation across source domains. As shown in Fig.~\ref{logits}, different specialist models yield varying predictions and corresponding confidences over the classes for the same image, thus the expert in one domain may be totally different from experts in other domains. Consequently, it becomes challenging  for a single model to effectively accommodate samples with various distributions to a uniform one-hot label and the one-hot labels are insufficient for learning diverse features.

Besides, considering that a model pre-trained on ImageNet~\cite{deng2009imagenet} is commonly chosen as the backbone in DG, the prevailing practice of assigning equal importance to different source domains is not always optimal for DG, because `ImageNet' would also introduce inductive bias to the pre-trained model~\cite{zhang2022towards}. As a result, domains with significant distribution deviation from `ImageNet' could dominate the optimization process, potentially hindering the learning from domains whose distributions are close to that of `ImageNet' or leading to catastrophic forgetting in those domains~\cite{gan2022decorate}. To address this issue, it is crucial to take into account that source domains with varying distribution distances from `ImageNet' could contribute differently to model updates. This necessitates the adoption of diverse training paces for different source domains to prevent overfitting in certain domains and catastrophic forgetting in others.

In this study, we theoretically and empirically analyze the drawback of employing one-hot labels in DG. Based on our findings, we propose a novel paradigm for enhancing generalization capacities, named Diverse Target and Contribution Scheduling~(DTCS), including two key modules: (1) Diverse Target Supervision~(DTS), which effectively addresses the issue of gradient conflicts during training at the source and therefore enables better convergence towards global optimum. (2) Diverse Contribution Balance~(DCB), which balances the distinct contributions of source domains. As shown in Fig.~\ref{archi}, we provide three distinctive paradigms for DTCS, utilizing different diverse target prophets, to emphasize that it is diverse targets rather than the model for generating them that matters. In summary, our main contributions are:

\begin{itemize}

   \item We theoretically and empirically elucidate the limitations inherent in employing one-hot labels as training targets in DG, and thereby bridge the relationship between distribution shifts and gradient conflicts for the first time. From a new perspective, we address the limitations of the commonly used one-hot labels and the equal weights for source domains in DG from the empirical source risk.
  
    \item We propose a novel paradigm for DG, named Diverse Target and Contribution Scheduling~(DTCS), which employs diverse targets across domains to alleviate gradient conflicts and dynamically trades off the contributions of different source domains.

    \item We conduct extensive experiments and analysis on four commonly-used benchmark datasets to verify the effectiveness and superiority of the proposed method against state-of-the-art DG approaches. Notably, the our proposed DTS could be easily plugged into existing DG methods with notable performance improvements.
\end{itemize}

\section{Related Work}

Numerous endeavours have been undertaken to enhance the generalization capabilities of hypothesis models. Mainstream approaches can be mainly categorized into three groups. 

(1) Increasing the diversity of source domains through data augmentation~\cite{volpi2018generalizing,shankar2018generalizing} or data generation~\cite{zhou2020domain,anoosheh2018combogan, xia2023generative}. DDAIG~\cite{zhou2020deep} augmented the training data of source domains with domain-adversarial image generation, thereby exposing more unseen data to the hypothesis model. CrossGrad~\cite{shankar2018generalizing} directly perturbed the inputs by adding the gradients of the label classifier and domain classifier to the original image. Considering the imbalanced data scale problem, GINet~\cite{xia2023generative} generated more novel samples for minority domain/category. 

(2) Learning domain-invariant representations through feature alignment strategies~\cite{ganin2016domain,li2018deep,zhao2020domain,yao2022pcl, wang2022contrastive}. DANN~\cite{ganin2016domain} introduced domain adversarial training to remove the domain-dependent signals, thereby extracting domain-independent features across domains.
PCL~\cite{yao2022pcl} and SelfReg~\cite{kim2021selfreg} took advantage of contrastive learning to encourage the samples' representations to be closer to those of the same class and further away from those of the other classes. CACE~\cite{wang2022contrastive} proposed ACE to quantify the causality of extracted features and then performed alignment across domains based on ACE. 

(3) Finding flat minima to lower the bound of target risk with ensemble techniques~\cite{cha2021swad, zhou2021domain}. SelfReg~\cite{kim2021selfreg} utilized stochastic weight averaging to optimize the loss space for flatter minima. SWAD~\cite{cha2021swad} proposed stochastic weight averaging densely to further bridge the flatness in loss landscapes to generalization performance. 

As another branch, Gradient Surgery~\cite{mansilla2021domain} introduced the perspective of viewing the scenarios of DG as a multi-task learning problem. It intuitively suggested the existence of gradient conflicts in DG and attempted to remove conflicting components considering only the signs of the gradient components. Furthermore, GradCa~\cite{song2023gradca} proposed `average' and `sign-mask' strategies to modify the gradient components, taking into account both their magnitudes and signs, avoiding the bad impact of only considering the gradient component signs on the total gradient updating. 

Different from existing studies addressing gradient conflicts~\cite{mansilla2021domain, song2023gradca, shi2022gradient} in DG, this paper offers two distinctive insights. (\romannumeral1) We provide empirical evidence and theoretical analysis to verify and explain why gradient conflicts occur in DG for the first time, introducing a novel perspective based on empirical source risk. (\romannumeral2) Based on this new view, we focus on alleviating gradient conflicts at the source level, diverging from existing methods~\cite{mansilla2021domain, song2023gradca, shi2022gradient} that manipulate conflicting components which can be seen as a domain-invariant learning strategy to maintain consistent gradients across domains.

\section{Methodology}

\begin{figure*}[t]
    \centering
    \subfloat[The framework of our proposed model\label{frame}]{\includegraphics[width=0.8\linewidth]{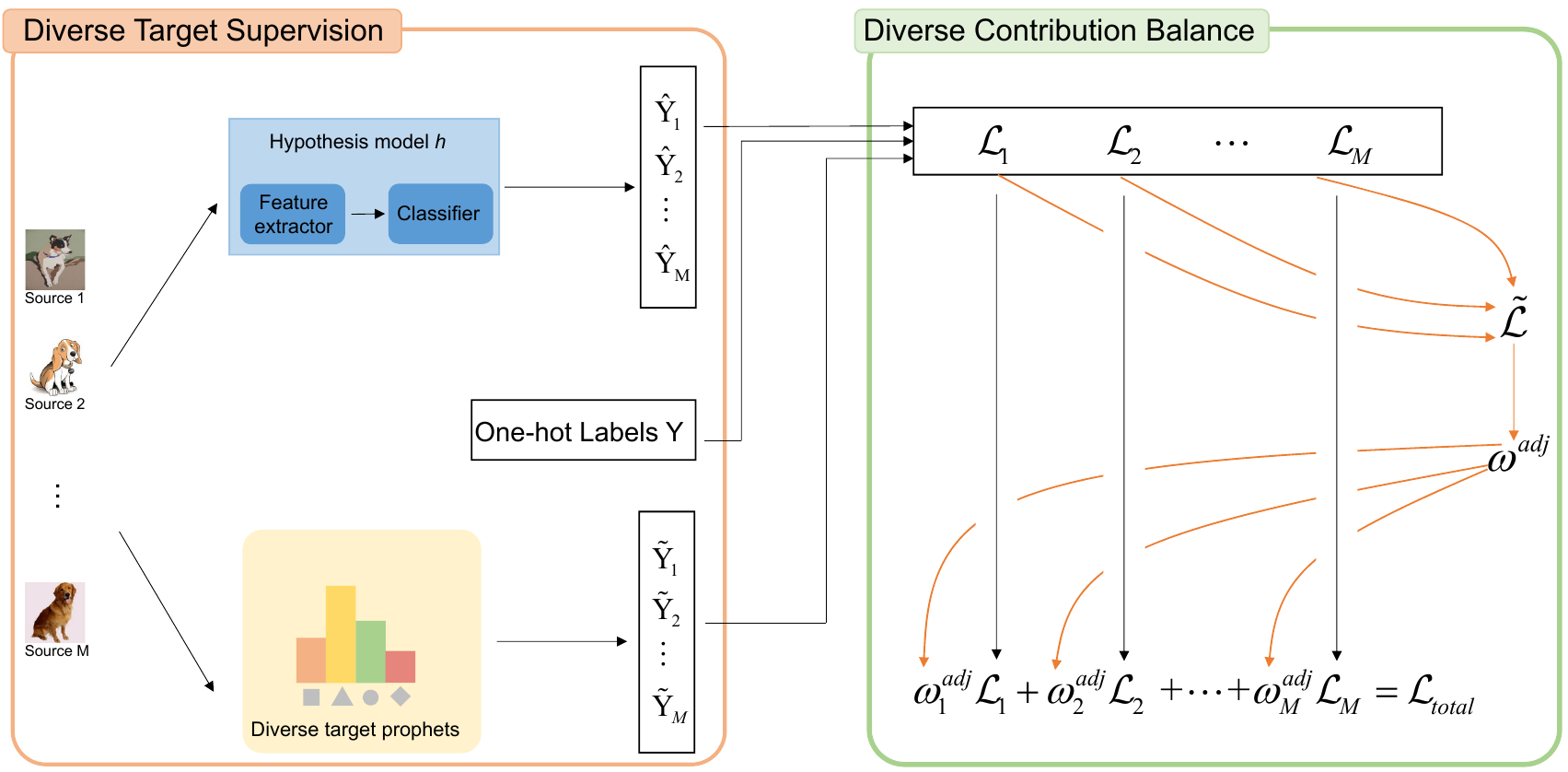}}
    
    \subfloat[The variants of diverse target supervision\label{variant}]{\includegraphics[width=0.8\linewidth]{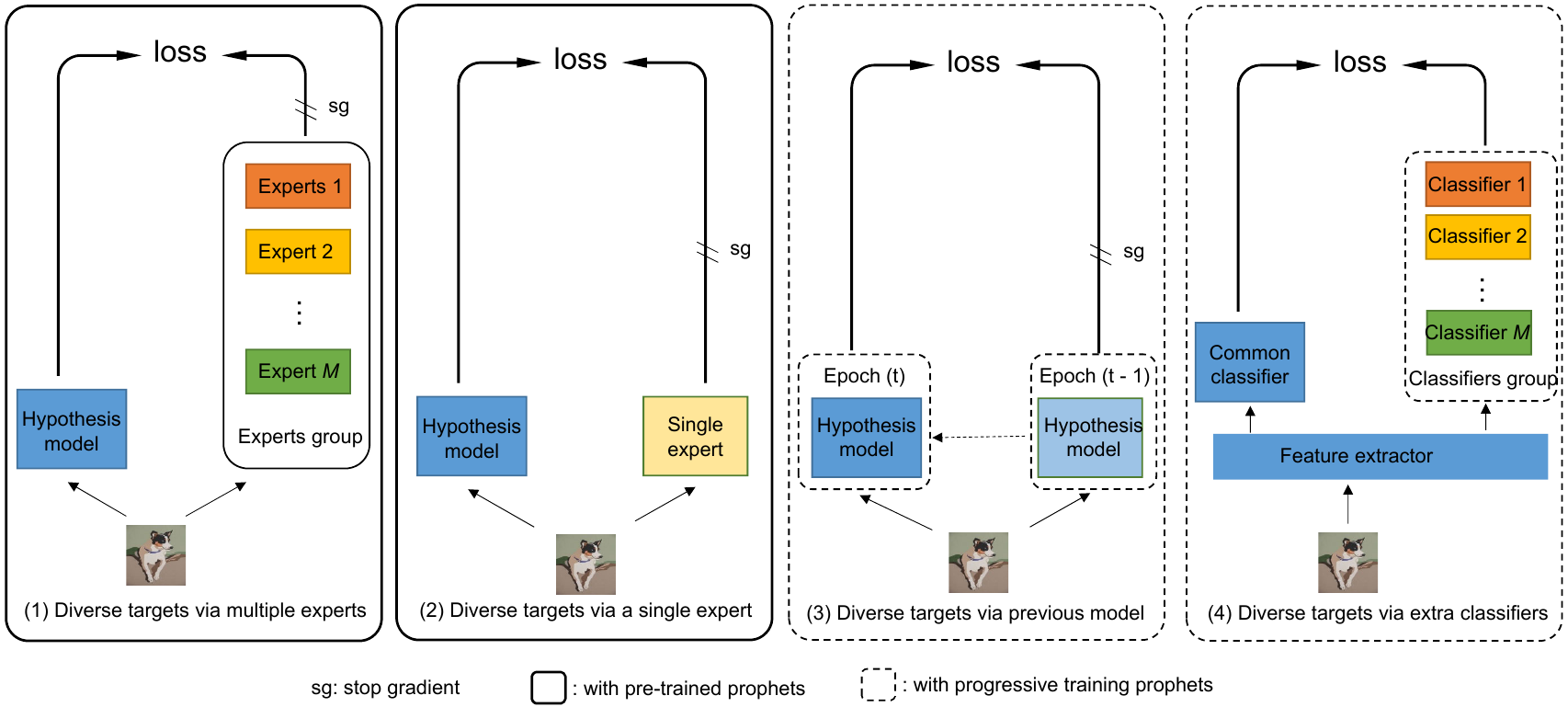}}

    \caption{The illustration of our proposed Diverse Target and Contribution Scheduling~(DTCS). (a) Addition to the one-hot labels, we also utilize the soft labels to surrogate the diverse targets, dividing the overall task into non-contradictory sub-tasks. Besides, we balance the contributions of different source domains via re-weighting their losses according to the relative inverse training rate every iteration instead of directly summing up or averaging the losses across source domains. (b) We provide four variants of diverse target supervision, with diverse target prophets ranging from complex models to simple models, from pre-trained models to progressive learning models, and from models with a large number of parameters to those with a reduced set.}
    
\end{figure*}

In this section, we begin by theoretically analyzing the conflicts in gradient descent directions from the perspective of empirical source risks, aiming to understand the underlying causes of these conflicts and their implications for the learning process. The analysis sheds light on a new direction for DG from the perspective of empirical source risk. Building upon this analysis, we propose a solution that assigns diverse targets for samples across source domains. By doing so, we effectively partition the overall optimization task into non-contradictory sub-tasks, which helps eliminate the negative effects of gradient conflicts. Additionally, we adaptively re-weight the effects of source domains, allowing us to balance their diverse influences. Fig.~\ref{frame} illustrates the overall Diverse Target and Contribution Scheduling~(DTCS) framework. Besides, to eliminate the influence of models which generate the diverse targets, we present four variants of diverse target prophets that range from complex models to simple models, which is shown in Fig.~\ref{variant}. 

\subsection{Preliminaries}
Let $x \in\mathcal{X}$ and $y \in\mathcal{Y}$ denote the input and output, respectively. $X$ and $Y$ represent the corresponding random variables. In DG, the dataset contains $N$ domains with different distributions: $D = \{D_i|i = 1, 2, \cdots, N\}$ where $D_i \sim \mathbb{P}^D_i(X, Y)$ and $\mathbb{P}^D_i(X, Y) \neq \mathbb{P}^D_j(X, Y), 1 \leq i \neq j \leq N$. Then those domains are partitioned into source domains $S = \{S_i|i = 1, 2, \cdots, M\}$ and target domains $T = \{T_i|i = 1, \cdots, N - M\}$, while the latter are inaccessible during training. The goal of DG is to learn a robust hypothesis model with minimal generalization risk on target domains: $\hat{h} = \arg \min_{h \in \mathcal{H}} \mathcal{E}_T(h) = \arg \min_{h \in \mathcal{H}} \frac{1}{N - M}\sum_{i = 1}^{N-M}\mathbb{E}_{(X, Y) \sim \mathbb{P}_i^T}[\ell (h(X), Y)]$, where $\mathcal{E}$ denotes the risk (error) of the hypothesis $h$, $\mathcal{H}$ is the hypothesis space and $\ell(\cdot, \cdot)$ is the loss function.

\subsection{Gradient Conflicts from the View of Empirical Source Risk}

Existing methods have achieved notable improvements in DG. However, gradient conflict is always a tricky problem that affects the generalization capacities. Specifically, gradient conflicts occur when reducing the empirical risk of one source domain interferes with the goal of minimizing the empirical risks of other source domains. This interference leads to loss fluctuations, as depicted in Fig.~\ref{ERM_loss}, and hampers the optimization process. Nonetheless, gaining a theoretical understanding of the origin and nature of gradient conflicts is crucial. To shed light on this issue, we present the following theorem which provides insights into the emergence of gradient conflicts based on the perspective of empirical source risk. 

\begin{theorem}
Consider a set of $M$ mini-batches $\{b_i\}_{i = 1}^{M}$ sampled from $M$ source domains such that the batch for training $B := \{b_i|i = 1, 2, \cdots, M\}$, $n = |b_1|= |b_2| = \cdots =|b_M|$ is the number of samples in one mini-batch. Let $\mathcal{H}$ be a hypothesis space of VC dimension $d$. If $h_{S_j}^* = \min_{h \in \mathcal{H}}\mathcal{E}_{S_j}(h)$ is the optimal error minimizer for mini-batch $b_j$, and $\hat{h} \in \mathcal{H}$ is the empirical minimizer of $\frac{1}{M - 1}\sum_{i \neq j} \hat{\mathcal{E}}_{S_i}$ over the remaining $M - 1$  mini-batches (excluding $b_j$). Then for any $\delta \in (0, 1)$, the following bound holds with probability at least $1-\delta$, 
\begin{equation}
\begin{aligned}
    \mathcal{E}_{S_j}(\hat{h}) \leq &\mathcal{E}_{S_j}(h_{S_j}^*) + 2\sqrt{\frac{8d\ln{\frac{2e(M-1)n}{d}} + 8\ln{\frac{4}{\delta}}}{(M-1)n}} \\
    & + \sum_{i \neq j}\frac{2}{M -1}{\rm IPM_G}(\mathbb{P}_i, \mathbb{P}_j), \label{bound}
\end{aligned}
\end{equation}
where ${\rm IPM_G}(\mathbb{P}_i, \mathbb{P}_j)= \sup\limits_{g \in {\rm G}}|\int_X g(x)(\mathbb{P}_i(x) - \mathbb{P}_j(x))\mathrm{d}x|$ is the Integral Probability Metric~\cite{muller1997integral}, measuring the discrepancy for distributions on the probability function over $X$, $\mathbb{P}_i$ and $\mathbb{P}_j$ are the distributions of source domain $i$ and $j$, respectively.
\label{theory_loss}
\end{theorem}

\textbf{Remark 1.} (Interpretations of Theorem~\ref{theory_loss}) In Theorem~\ref{theory_loss}, the risk of the source domain $j$ is bounded by three terms: (1) the optimal risk $\mathcal{E}_{S_j}(h_{S_j}^*)$, (2) a bound that depends on the number of training samples, and (3) the distribution gap between the source domain $j$ and the other source domains. Consequently, during the training process, the empirical risk for the source domain $j$ may not reach its optimum due to the inherent distribution shift among source domains. Alternatively, if the empirical loss for the source domain $j$ is minimized, the hypothesis model may excessively focus on this particular domain, leading to a sub-optimal solution and potentially disregarding valuable knowledge from the remaining source domains in the current epoch. Conversely, when the hypothesis model shifts its attention to other source domains in the subsequent epoch, the empirical risk of the source domain $j$ may increase once again. This phenomenon manifests as the observed fluctuations in loss values, as illustrated in Fig.~\ref{ERM_loss}, providing the evidence of the existence of gradient conflicts during the training process. We provide the detailed proof of Theorem~\ref{theory_loss} in the Supplemental Material.

As is known, the target risk in DG is up-bounded by three factors: the empirical source risk, the bound associated with hypothesis space, and the distribution discrepancy between source domains and target domains~\cite{ben2010theory}. However, due to the presence of gradient conflicts, it becomes challenging for the empirical risks of source domains to reach their respective optimum simultaneously. Consequently, the hypothesis model inadequately learns from source domains, leading to an increase in the upper bound of the target risk. As a result, the hypothesis model suffers in generalizing out of samples.

\subsection{Diverse Target Supervision~(DTS)}

Given the varying distributions of the same class across source domains, employing identical one-hot labels as the training targets may have a detrimental effect on the models' generalization performance. This is primarily attributed to the presence of distribution shift, which could lead to conflicts in the gradient descent direction during training. Meanwhile, Fig.~\ref{logits} points out that different domains have different specialist experts in the task of image recognition, and a single expert may not perform optimally in novel environments due to the distribution shift. This observation underscores the inherent challenge of effectively learning from multiple source domains using solely the one-hot labels. 

To mitigate the adverse effect of gradient conflicts and effectively learn the knowledge in source domains, we propose to adopt diverse targets that correspond to different modalities of the distribution, as opposed to using uniform one-hot labels. The diverse targets aim to implicitly partition the data pertaining to a specific class across domains into sub-datasets with distinct training targets. By doing so, we divide the overall task into non-contradictory sub-tasks, effectively addressing the issue of gradient conflicts arising from distribution shift and thereby facilitating the optimization process. It is worth noting that although the diverse targets for a given class may differ, they still encapsulate the same underlying class information, signifying the same class. 

Moreover, considering the potential benefits of augmenting the source domains for improving generalization performance\cite{volpi2018generalizing,shankar2018generalizing}, the introduction of diverse targets bearing different extra information can help to capture underlying semantics and thereby implicitly diversify the intra-class variations. 

In practice, we adopt the soft labels, which are predictions for the samples, as surrogates for the diverse targets. Soft labels provide a means to capture other semantic information that is discarded by one-hot labels and can vary across domains for the same class, as illustrated in Fig.~\ref{logits}. By employing soft labels as supervision, the hypothesis model is able to attend to various information that is not conveyed in one-hot labels, and thereby accommodate various features across domains. As a result, the optimization objective can be formulated as:
\begin{equation}
    \min \mathcal{L}_{total} = \min \sum_{i = 1} ^{M} \mathcal{L}_i, \label{kd_loss}
  \end{equation}%
with 
\begin{align}
\hspace{-0.3em}\mathcal{L}_i = \alpha CE_{\mathbb{P}_i^S}(h(X), Y) + (1 - \alpha)\tau^2 KL_{\mathbb{P}_i^S}(h^{\tau}(X), \tilde{Y}^{\tau}), \label{single_loss}
\end{align}%
where $\alpha$ is the weighting factor, $\tau$ is the temperature for controlling the softness of the probability over classes, $Y$ is the one-hot label, and $\tilde{Y}$ is the diverse target. The objective comprises two terms: the first term is the cross entropy between the hypothesis model's outputs and the one-hot labels ($CE$ denotes the Cross-Entropy), while the second term is the KL divergence between the diverse targets and the hypothesis model's outputs ($KL$ denotes the KL Divergence). The $m$-th element of $\tilde{Y}^{\tau}$ and $h^{\tau}(X)$ are defined as:
\begin{align}
   \tilde{Y}^{\tau}_m = \frac{\exp(\tilde{z}_m / \tau)}{\sum_n \exp(\tilde{z}_n/ \tau)}, \qquad  h_{m}^{\tau}(X) = \frac{\exp(\hat{z}_m / \tau)}{\sum_n \exp(\hat{z}_n/ \tau)}, \label{stu}
\end{align}%
where $\tilde{z}$ and $\hat{z}$ denote the diverse target and the hypothesis model's output, respectively, and the subscript $m$ represents their $m$-th element. The transformations in Eq.~\ref{stu} degrade to the vanilla softmax function when $\tau = 1$.

Next, we present four distinct approaches for generating soft labels to serve as surrogates for the diverse targets. These methods encompass pre-trained multiple experts, pre-trained single expert, the hypothesis model in previous epoch and multiple linear classifiers. Notably, these approaches vary in terms of model complexity, whether they are pre-trained or not, and parameter quantity. They span from complex models to simple ones, from pre-trained models to progressive learning models, and from models with a large number of parameters to those with a reduced set.

\subsubsection{Diverse Targets through Multiple Experts}
We leverage specialist experts specific to each domain to generate soft labels for samples within their respective domains. To achieve this, we first train multiple specialist experts, each tailored to a specific source domain. During the training of the hypothesis model, when learning samples from a particular domain, we utilize the corresponding expert for that domain to generate the soft labels. It is important to note that the experts do not require backpropagation for updates at this stage. Denote the pre-trianed experts as $\{E_i\}_{i = 1}^{M}$, where $E_i$ corresponds to source domain $i$. These experts take the original image as input and produce the soft label as output:
\begin{equation}
 E_i =  \arg \min_{h \in \mathcal{H}} \mathbb{E}_{(X, Y) \sim \mathbb{P}_i^S}[\ell (h(X), Y)], 
    \tilde{Y} = E_{i}(X).
    \label{ME}
\end{equation}
\subsubsection{Diverse Targets through a Single Expert}
In order to mitigate the cost associated with multiple experts, we also explore the use of a single expert pre-trained on all source domains to generate soft labels. To accomplish this, we first train an expert using samples from all source domains. During training for the hypothesis model, the differentiation between source domains is no longer necessary. Instead, we employ this single expert to produce soft labels for samples across all source domains. Similar to the previous approach, no backpropagation is required for this expert at this stage. Denote the pre-trained single expert as $E$, whose input is the original image and output is the soft label:
\begin{equation}
\hspace{-0.7em}E =  \arg \min_{h \in \mathcal{H}} \frac{1}{M}\sum_{i = 1}^{M}\mathbb{E}_{(X, Y) \sim \mathbb{P}_i^S}[\ell (h(X), Y)], 
    \tilde{Y} = E(X).
    \label{SE}
\end{equation}
\subsubsection{Diverse Targets through Hypothesis Model of Previous Epoch}
To save the computational cost associated with pre-training, we also explore an alternative approach that takes advantage of the progressive training hypothesis model for generating soft labels. However, since the predicted soft labels by the current model do not provide meaningful supervision, we make use of the model state in the previous epoch to generate the soft labels. More specifically, let $h_{t}$ and $h_{t - 1}$ represent the current hypothesis model and the model of the previous epoch, respectively. The previous model, $h_{t - 1}$, also does not require backpropagation in the current epoch, takes the original image as input, and then produces the soft label as output during the training of the hypothesis model:
\begin{equation}
    \tilde{Y} = h_{t - 1}(X),
    \label{ONSE}
\end{equation}
with the relation between $h_t$ and $h_{t - 1}$:
\begin{equation}
    \theta_{h_{t}} = \theta_{h_{t - 1}} - \frac{\eta}{M}\nabla(\sum_{i = 1}^{M}\mathbb{E}_{(X, Y) \sim \mathbb{P}_i^S}[\ell (h_{t - 1}(X), Y)])), 
\end{equation}
where $\theta$ and $\eta$ denote the hypothesis model's parameters and step size, respectively.
\subsubsection{Diverse Targets through Multiple Linear Classifiers}
In order to reduce the cost associated with model size, we explore a simplified method that just adopts multiple linear classifiers to generate soft labels. These linear classifiers, each tailored to a specific domain, are all built upon a same feature extractor. No additional pre-training is required for these linear classifiers, and we only update the classifiers specific to each domain within their respective domain along with the shared feature extractor. Denote these linear classifiers and feature extractor as $\{g_i\}_{i = 1}^M$ and $f$, respectively, where $g_i$ corresponds to the source domain $i$. These linear classifiers take the features from the feature extractor as input and produce the corresponding soft labels:
\begin{equation}
\hspace{-0.4em}g_i = \arg \min_{g \in \mathcal{G}}\mathbb{E}_{(X, Y) \sim \mathbb{P}_i^S}[\ell (g(f(X)), Y)],
    \tilde{Y} = g_i(f(X)),
    \label{MC}
\end{equation}
where $\mathcal{G}$ is the hypothesis space of these linear classifiers.

\subsection{Diverse Contribution Balance~(DCB)}
The ImageNet~\cite{deng2009imagenet} pre-trained ConvNet is the most common choice as the backbone network for feature extraction or logit prediction in DG. The pre-trained backbone is then fine-tuned using samples from source domains. This manner may overlook a crucial aspect: ImageNet itself can be regraded as a novel domain with distinct distribution~\cite{zhang2022towards}. Consequently, domains whose distributions deviate significantly from that of ImageNet may dominate the process of updating the model, potentially hindering the learning from domains whose distributions are close to that of ImageNet or leading to catastrophic forgetting in those domains~\cite{gan2022decorate}. 

However, most mainstream solutions in DG tend to overlook the diverse contributions of source domains and equally treat the source domains when updating the hypothesis model. Recognizing the importance of differentiating between source domains, SelfReg~\cite{kim2021selfreg} designed a curriculum learning strategy to sequentially learn from source domains according to their distribution distances from the dataset on which the backbone is pre-trained. DARLING~\cite{zhang2022towards} highlighted the importance of pre-training protocol on heterogeneous data and alternated the ImageNet pre-trained protocol with a domain-aware one.

To fairly learn from diverse source domains, it is essential to acknowledge the distinctions between source domains and balance their respective effects. As a result, we attempt to strike a harmonious trade-off among source domains by dynamically balancing their contributions at every iteration. The optimization objective at iteration $n$ can be formulated as:
\begin{algorithm}[tb]
\caption{Training algorithm for DG via DTCS}
\label{algorithm}
\textbf{Input}: $M$ source domains: $\{S_i\}_{i = 1}^{M}$, diverse target prophets\\
\textbf{Parameter}: Weighting factor: $\alpha$, momentum coefficient: $m$, temperature: $\tau$\\
\textbf{Output}: the hypothesis model $h$
\begin{algorithmic}[1] 
\WHILE{training is not converged}
\FOR{$i = 1$ to $M$}
\STATE Sample data from $S_i$
\STATE Calculate the soft labels with the diverse target prophets choosing from Eq.~\ref{ME}, Eq.~\ref{SE}, Eq.~\ref{ONSE} or Eq.~\ref{MC}
\STATE Calculate the domain-specific loss by Eq.~\ref{single_loss}
\STATE Update the domain-specific weight by Eq.~\ref{weight}
\ENDFOR
\STATE Calculate the re-weighted total loss by Eq.~\ref{weight_loss}
\STATE Update the hypothesis model $h$ by minimizing the re-weighted total loss 
\ENDWHILE
\end{algorithmic}
\end{algorithm}
\begin{align}
\mathcal{L}_{total} =\sum_{i = 1} ^{M} \omega_i(n) \mathcal{L}_i(n), \label{weight_loss}
\end{align}%
where $\omega_i(n)$ is the weight for balancing the effect of the source domain $i$ at iteration $n$, $\mathcal{L}_i(n)$ is the $i$-th domain's loss at iteration $n$ and $\mathcal{L}_i$ is defined in Eq.~\ref{single_loss}. Through assigning different importances to source domains, we can control the influences of different source domains. The goal of the domain balance strategy is to dynamically adjust the learning pace within each domain, ensuring that the loss for each source domain can fairly drop and avoiding the issue that a subset of source domains dominate the training process and thereby impede the learning from other domains.

\textbf{Momentum update for weights.} Inspired by Gradient normalization~\cite{chen2018gradnorm} and Dynamic Weight Averaging~\cite{liu2019end} in multi-task learning, we adopt the inverse training rate $\tilde{\mathcal{L}}_i(n)= \mathcal{L}_i(n) / \mathcal{L}_i(n-1)$ to reflect the decline of the $i$-th domain loss ($\mathcal{L}_i(n-1)$ is the loss of source domain $i$ in previous iteration), and then adjust the $i$-th domain's weight according to the relative inverse training rate $\omega^{adj}_i = \tilde{\mathcal{L}}_i(n) / \sum_j \tilde{\mathcal{L}}_j(n)$. 

To make the weights evolve smoothly, we propose a momentum update for the weights. Formally, the weight for domain $i$ at iteration $n$ can be updated by:
\begin{align}
   \omega_i(n) \leftarrow  m\omega^{adj}_i + (1 - m)\omega_i(n - 1), \label{weight}
\end{align}%
where $m \in [0, 1]$ is the momentum coefficient. The momentum update for the weights of source domains dynamically balances the contributions of different domains according to their respective paces of gradient descent. This strategy guarantees a fair drop in losses across source domains, preventing the model from excessively focusing on a subset of source domains. 

The pseudo-code for our algorithm is shown in Algorithm~\ref{algorithm}.

\begin{table}
    \centering
    \setlength{\tabcolsep}{4pt}
    \caption{Results on PACS with object recognition accuracies (\%). Higher is better, bold indicates the best performance out of all compared methods. $\dag$: without ensembling technique.}
    \begin{tabular}{l|cccc|c}
    \toprule
    \multirow{2}{*}{Method} & \multicolumn{4}{c|}{Target domain} & \multirow{2}{*}{Avg($\uparrow$)}\\
    \cmidrule(r){2-5}
     & A & C & P & S \\
    \hline
    \midrule
    IRM~\cite{arjovsky2019invariant}       & $79.26$         & $75.32$        & $94.24$         & $71.72$        & $80.13$ \\
    GroupDRO~\cite{sagawa2019distributionally}  & $77.73$         & $74.89$        & 95.66           & $73.76$        & $80.51$   \\
    Mixup~\cite{yan2020improve}     & $79.38$         & 78.79          & $94.39$         & $68.86$        & $80.35$  \\
    MMD~\cite{li2018domain}       & $77.79$         & $71.43$        & $94.31$         & $73.73$        & $79.32$ \\
    VREx~\cite{krueger2021out}      & $80.84$         & $70.95$        & $93.64$         & 78.44          & $80.97$ \\
    RSC~\cite{huang2020self}       & 79.88         & 76.87        & 94.56         & 77.11        & 82.10 \\
    DANN~\cite{ganin2016domain}      & $77.43$         & $66.84$        & \bfseries96.63  & $69.82$        & $77.68$ \\
    CDANN~\cite{li2018deep}     & $75.17$         & 73.99          & $95.66$         & $72.93$        & $79.44$ \\
    MTL~\cite{blanchard2021domain}       & $79.99$         & $72.18$        & $95.28$         & $74.94$        & $80.60$  \\
    SagNet~\cite{nam2021reducing}    & 81.15           & $75.05$        & $94.61$         & 75.38          & $81.55$ \\
    ARM~\cite{zhang2021adaptive}       & $80.42$         & $75.96$        & $95.21$         & $72.33$        & $80.98$ \\
    Fish~\cite{shi2022gradient}            & 80.70           & 74.52          & 95.24          & 75.21 & 81.41\\
    GradCa~\cite{song2023gradca}    & $82.51$         & $78.66$        & $96.47$         & $77.25$        & $83.72$ \\
    SWAD~\cite{cha2021swad}      & $83.28$         & $74.63$        & $96.56$         & $77.96$        & $83.11$ \\
    {PCL\dag}~\cite{yao2022pcl}  & \bfseries83.71  & 72.18          & 94.61           & 71.06          & $80.39$\\
    PCL~\cite{yao2022pcl}       & 83.53           & 73.61          & $96.18$         & $77.20$        & $82.63$\\
    \hline
    {Ours~(ME)} & 81.15           & \bfseries78.92 & $93.95$         & \bfseries81.73 & \bfseries83.94 \\
    {Ours~(SE)} & 82.28           & 78.11          & $94.61$         & 78.26          & 83.31 \\
    {Ours~(MP)}& 82.28          & 78.24          & $93.53$         & 78.42          & 83.12 \\
    {Ours~(MC)} & 81.59           & 78.46          & $94.49$         & 79.43          & 83.49 \\

    \bottomrule
    \end{tabular}
    
    \label{PACS}
\end{table}

\section{Experiment}\label{exp}

To evaluate the effectiveness and advantages of the proposed DTCS, we conduct experiments on four benchmarks: PACS~\cite{li2017deeper}, Office-Home~\cite{venkateswara2017deep}, Terra-Incognita~\cite{beery2018recognition}, and VLCS~\cite{fang2013unbiased}. We also compare the performance of our DTCS with state-of-the-art approaches in DG, including IRM~\cite{arjovsky2019invariant}, GroupDRO~\cite{sagawa2019distributionally}, Mixup~\cite{yan2020improve}, MMD~\cite{li2018domain}, VREx~\cite{krueger2021out}, RSC~\cite{huang2020self}, DANN~\cite{ganin2016domain}, CDANN~\cite{li2018deep}, MTL~\cite{blanchard2021domain}, SagNet~\cite{nam2021reducing}, ARM~\cite{zhang2021adaptive}, Fish~\cite{shi2022gradient}, GradCa~\cite{song2023gradca}, SWAD~\cite{cha2021swad} and PCL~\cite{yao2022pcl}.

\subsection{Benchmarks}
(1) \textbf{PACS} is composed of 9991 images categorized into 7 classes from four domains: Art painting (A), Cartoon (C), Photo (P) and Sketch (S). (2) \textbf{Office-Home} is designed for domain adaptation, consisting of 15588 images distributed across 5 categories and four styles: Art (A), Clipart (C), Product (P) and Real-World (R). (3) \textbf{Terra-Incognita} comprises wildlife photographs captured across different locations, namely L100, L38, L43, and L46. It consists of 24330 images distributed in 10 classes. (4) \textbf{VLCS} contains four datasets: Caltech101 (C), LabelMe (L), SUN09 (S), PASCAL VOC2007~(V), including 10729 images distributed in 5 categories.

\begin{table}
    \centering
    \setlength{\tabcolsep}{4pt}
    
    \caption{Results on Office-Home with object recognition accuracies~(\%). Higher is better, the best result is highlighted in bold. $\dag$: without ensembling technique.}
    \begin{tabular}{l|cccc|c}
    \toprule
    \multirow{2}{*}{Method} & \multicolumn{4}{c|}{Target domain} & \multirow{2}{*}{Avg($\uparrow$)}\\
    \cmidrule(r){2-5}
     & A & C & P & R \\
    \hline
    \midrule
    IRM~\cite{arjovsky2019invariant}        & $50.31$         & $46.94$        & $66.81$         & $69.94$         & $58.50$\\
    GroupDRO~\cite{sagawa2019distributionally}   & $56.69$         & $46.79$        & 71.17           & $71.31$         & $61.49$  \\
    Mixup~\cite{yan2020improve}      & $55.41$         & 49.86          & 72.10           & 74.33           & $62.92$ \\
    MMD~\cite{li2018domain}        & $54.48$         & 49.94          & $68.16$         & $72.52$         & $61.28$\\
    VREx~\cite{krueger2021out}       & $49.43$         & $45.82$        & $68.07$         & 68.19           & $57.88$\\
    RSC~\cite{huang2020self}        & $49.38$         & $45.91$        & $66.84$         & $67.41$         & $57.38$ \\
    DANN~\cite{ganin2016domain}       & $50.10$         & $48.05$        & 67.03           & $69.94$         & $58.78$ \\
    
    CDANN~\cite{li2018deep}      & $49.90$         & 46.62          & $68.36$         & $70.97$         & $58.96$ \\
    MTL~\cite{blanchard2021domain}        & $52.58$         & $46.99$        & $70.83$         & $72.46$         & $60.72$  \\
    SagNet~\cite{nam2021reducing}     & 56.28           & 51.32          & $70.64$         & 73.38           & $62.90$\\
    ARM~\cite{zhang2021adaptive}        & $52.68$         & $45.82$        & $68.64$         & $71.40$         & $59.63$ \\
    Fish~\cite{shi2022gradient}            & 54.69           & 48.54          & 70.97           & 72.83 & 61.76\\
    GradCa~\cite{song2023gradca}     & $57.88$         & $49.35$        & $72.91$         & \bfseries75.89  & $64.01$ \\
    SWAD~\cite{cha2021swad}       & $54.33$         & $49.80$        & $70.92$         & $71.97$         & $61.75$ \\
    {PCL\dag}~\cite{yao2022pcl}   & 57.36           & 50.23          & 71.62           & 73.32           & $63.14$\\
    PCL~\cite{yao2022pcl}        & $56.69$         & \bfseries52.49 & $72.24$         & $74.50$         & $63.98$\\
   
    \hline
    {Ours~(ME) } & \bfseries58.43  & 51.91          & \bfseries73.12  & 74.89           & \bfseries64.59\\
    {Ours~(SE)}  & 58.47           & 50.63          & 72.04           & 75.67           & 64.20 \\
    {Ours~(MP)}& 56.86           & 51.98          & 71.93           & 75.03           & 63.95 \\
    {Ours~(MC)}  &57.52            &49.51           & 72.49           & 74.18           &63.42  \\

    \bottomrule
    \end{tabular}
    \label{office_home}
\end{table}

\subsection{Implementation Details}
For (1) diverse targets generated by multiple experts~(ME), we employ ResNet-50~\cite{he2016deep} pre-trained on ImageNet as the backbone for these extra experts. Each expert is fine-tuned individually using the corresponding source domain. (2) Diverse targets generated by a single expert~(SE), we utilize ResNet-18 pre-trained on ImageNet as the backbone. The single expert is fine-tuned using all the source domains. Those pre-trained experts in~(1) and~(2) are trained using Adam optimizer for 60 epochs, and the learning rate is initialized at 5e-5 and decayed by 0.1 at 70\% and 90\% of the total epochs. The weight-decay and batch size are fixed as 5e-4 and 32, respectively. (3) Diverse targets generated by the hypothesis model at the previous epoch~(MP), there is no need for an extra architecture or pre-training, we simply reload the model state from the previous epoch and save the model state after every epoch. (4)~Diverse targets generated by multiple linear classifiers~(MC), the extra linear classifiers are composed of just one fully connected ~(FC) layer. We train those extra classifiers simultaneously with the hypothesis model using the same optimization strategy.

For the hypothesis model, we employ ResNet-18 pre-trained on ImageNet as the backbone. We train the hypothesis model for 5k iterations using SGD optimizer, with the learning rate decayed by 0.1 at 60\% and 80\% of the total epoch. The weight-decay and batch size for each domain are 5e-4 and 32, respectively. In order to ensure domain balance, the weight for each source domain is initialed at $1/M$, and $m$ is decayed by 0.1 at 60\% and 80\% of the total epoch. We provide the search space for other hyperparameters in TABLE~\ref{parameter}.  

\begin{table}[h]
    \centering
    \caption{Hyperparameter search space.}
    \begin{tabular}{ccc}
    \toprule
         Parameter & Search space \\
         \midrule
         learning rate & [5e-3, 3e-3, 1e-3, 5e-4] \\
         temperature $\tau$ & [0.5, 1, 2, 5] \\
         weighting factor $\alpha$ & [0.1, 0.2, 0.5]\\
         momentum coefficient $m$ & [0.9, 1.0]\\
         \bottomrule
    \end{tabular}
    \label{parameter}
\end{table}

\begin{table}
    \centering
    \setlength{\tabcolsep}{4pt} 
    \caption{Results on Terra-Incognita with object recognition accuracies (\%). Higher is better,the best result is highlighted in bold. $\dag$: without ensembling technique.}
    \begin{tabular}{l|cccc|c}
    \toprule
    \multirow{2}{*}{Method} & \multicolumn{4}{c|}{Target domain} & \multirow{2}{*}{Avg($\uparrow$)}\\
    \cmidrule(r){2-5}
     & L100 & L38 & L43 & L46 \\
    \hline
    \midrule
   IRM~\cite{arjovsky2019invariant}       & $51.33$         & $31.06$        & $50.38$         & $33.46$        & $41.56$ \\
    GroupDRO~\cite{sagawa2019distributionally} & \bfseries54.31  & $34.95$        & 52.02           & $33.33$        & 43.65  \\
    Mixup~\cite{yan2020improve}    & $49.96$         & 33.38          & 51.29           & 30.23          & $41.22$\\
    MMD~\cite{li2018domain}      & $49.96$         & 19.94          & $51.04$         & $27.70$        & $37.16$ \\
    VREx~\cite{krueger2021out}     & $40.65$         & $29.95$        & $50.06$         & 33.72          & $38.60$\\
    RSC~\cite{huang2020self}      & $47.32$         & $37.66$        & $51.67$         & 35.95          & $43.15$\\
    DANN~\cite{ganin2016domain}     & $32.38$         & $23.55$        & 44.36           & $31.40$        & $32.92$  \\
    
    CDANN~\cite{li2018deep}    & $40.23$         & 31.04          & $43.01$         & $33.84$        & $37.03$\\
    MTL~\cite{blanchard2021domain}      & $38.94$         & $35.18$        & $52.80$         & $35.29$        & $40.55$ \\
    SagNet~\cite{nam2021reducing}   & 47.25           & 29.67          & 52.87           & 25.22          & $38.75$\\
    ARM~\cite{zhang2021adaptive}      & $44.98$         & $33.73$        & $43.39$         & $27.77$        & $37.47$\\
    Fish~\cite{shi2022gradient}            & 45.74           & 29.30          & 54.72           & 33.76 & 40.88\\
    SWAD~\cite{cha2021swad}     & $49.80$         & $33.16$        & \bfseries55.57  & $33.19$        & $42.93$\\
    {PCL\dag}~\cite{yao2022pcl} & 45.74           & 33.28          & 51.17           & 32.10          & $40.57$\\
    PCL~\cite{yao2022pcl}      & $52.62$         & $39.98$        & $48.49$         & $31.74$        & $43.21$\\
    
    \hline
    {Ours~(ME) }& 48.64          & 40.48          & $52.82$         & 36.39          & 44.58 \\
    {Ours~(SE)} & 49.02          & 43.07          &  53.38          & \bfseries36.82 & \bfseries45.57 \\
    {Ours~(MP)}& 48.83         & 42.26          & \bfseries56.80  & 33.57          & 45.36 \\
    {Ours~(MC)} &46.99           &\bfseries44.60  & 48.46           &  36.75         & 44.20  \\

    \bottomrule
    \end{tabular}
    \label{terra}
\end{table}

For a fair comparison, we adhere to the evaluation protocol in DomainBed~\cite{gulrajani2020search} by adopting the training-domain validation, choosing one domain as the target domain, and training on the remaining source domains. We split each source domain with 80\%/20\% train/validation ratio, and the validation parts in all source domains together form the validation set for model selection and evaluation.

\subsection{Experimental Results on Benchmarks}

TABLE~\ref{PACS} reports the object recognition accuracy on PACS. The proposed method demonstrates superior average recognition accuracy compared to other approaches across different scenarios. Notably, our method exhibits substantial improvements in the worst-case scenarios where the target domain is either `Sketch' or `Cartoon'. This highlights the necessity of diverse targets for mitigating gradient conflicts and domain balance for overcoming the issue of the training process being dominated by a subset of source domains.

As represented in TABLE~\ref{office_home}, the highest accuracy of average recognition accuracy on Office-Home indicates the superiority of the proposed approach. Moreover, our method performs comparably to the SOTA in the scenario where the target domain is either `Clipart' or `Real-World'. These findings demonstrate that the proposed modules effectively supervise the hypothesis model, leading to better performance.

TABLE~\ref{terra} summarizes the performance on Terra-Incognita, and showcases the highest average object recognition accuracy achieved by our proposed strategy. Notably, our method outperforms other approaches in the worst-case scenarios where either `L38' or `L46' is the target domain. This provides empirical evidence supporting the advantages of diverse target supervision and diverse contribution balance.

Results on VLCS are reported in Table~\ref{VLCS}. While our proposed method may not surpass the SOTA in individual scenarios, it is important to note that the scene-centric nature of images in VLCS poses a greater challenge for generalization compared to object-centric images~\cite{zhao2020domain}. Nonetheless, the highest average accuracy achieved by our method still verifies its effectiveness in this challenging task.

\begin{table}
    \centering
    \setlength{\tabcolsep}{4pt}
    
    \caption{Results on VLCS with object recognition accuracies (\%). Higher is better, bold indicates the best performance out of all compared methods. $\dag$: without ensembling technique.}
    \begin{tabular}{l|cccc|c}
    \toprule
    \multirow{2}{*}{Method} & \multicolumn{4}{c|}{Target domain} & \multirow{2}{*}{Avg($\uparrow$)}\\
    \cmidrule(r){2-5}
     & C & L & S & P \\
    \hline
    \midrule
    IRM~\cite{arjovsky2019invariant}      & $96.82$         & \bfseries67.29 & $65.58$         & $73.34$        & $75.76$ \\
    GroupDRO~\cite{sagawa2019distributionally} & $97.09$         & $59.77$        & 68.89           & $71.83$        & $74.39$   \\
    Mixup~\cite{yan2020improve}    & \bfseries98.32  & 64.80          & $69.35$         & $70.86$        & $75.83$ \\
    MMD~\cite{li2018domain}      & $97.88$         & $64.28$        & $67.10$         & 76.16          & $76.35$  \\
    VREx~\cite{krueger2021out}     & $96.20$         & $62.97$        & \bfseries73.65  & 73.68          &  $76.62$\\
    RSC~\cite{huang2020self}      & $93.29$         & $64.47$        & $71.52$         & $73.31$        & $75.65$\\
    DANN~\cite{ganin2016domain}     & $97.88$         & $59.53$        &69.00            & $74.79$        & $75.30$ \\
    
   CDANN~\cite{li2018deep}    & $96.29$         & 63.58          & $68.62$         & 76.66          & 76.29\\
    MTL~\cite{blanchard2021domain}      & $96.38$         & $62.54$        & $70.91$         & $71.68$        & $75.38$ \\
    SagNet~\cite{nam2021reducing}   & 97.09           & $62.07$        & $70.37$         & 75.42          & 76.24\\
    ARM~\cite{zhang2021adaptive}      & $96.29$         & $61.55$        & $72.32$         & $76.27$        & $76.61$\\
    
    Fish~\cite{shi2022gradient}            & 96.56           & 62.21          & 70.83           & 75.38 & 76.24\\
    SWAD~\cite{cha2021swad}            & 97.70           & 61.27          & 70.72           & \bfseries76.71 & 76.60\\
    {PCL\dag}~\cite{yao2022pcl}        & 95.50           & 62.07          & 71.40           & 74.49          & 75.86\\
    PCL~\cite{yao2022pcl}             & 97.09           & 62.07          & 71.06           & 75.05          & 76.32\\
    
    \hline
    {Ours~(ME)}       & 97.95           & 64.95          & 69.93           & 74.56          & 76.85 \\
    {Ours (SE) }      & 96.40           & 65.29          & 71.45           & 72.78          & 76.48 \\
    {Ours (MP) }      & 96.89           & 65.10          & 71.66           & 75.30          & \bfseries77.24 \\
    {Ours (MC) }      & 96.68           &64.72           &72.82            &73.49           &76.93  \\

    \bottomrule
    \end{tabular}
    \label{VLCS}
\end{table}

\section{Empirical Analysis}
In this section, we conduct ablation studies to investigate the impact of DTS and DCB, and analyze why they work. These studies aim to provide insights into the effectiveness and roles of components within our proposed method.

\noindent\textbf{Ablation Study.} To gain a comprehensive understanding of the contributions made by DTS and DCB, we conduct experiments on PACS by removing specific components from the proposed model. The baseline represents the DeepAll method, which refers to the vanilla network without any generalization techniques. As shown in TABLE~\ref{componets}, introducing diverse targets for source domains leads to significant improvement in generalization performance, with an increase of 6.66\% (C compared to A). Besides, DCB proves to be beneficial for achieving enhanced performance, as observed in the comparisons between A and B, as well as C and D.

\begin{table}[t]
    \centering
    \setlength{\tabcolsep}{2pt}
    
    \caption{Ablation study by removing different components from the proposed model on PACS. DCB denotes diverse contribution balance and DTS denotes diverse target supervision.}
    \begin{tabular}{l|c|cc|cccc|c}
    \toprule
    \multicolumn{2}{c|}{\multirow{2}{*}{Model}}& \multicolumn{2}{c|}{Component} & \multicolumn{4}{c|}{Target domain} & \multirow{2}{*}{Avg($\uparrow$)} \\
    \cmidrule(r){3-4}\cmidrule(r){5-8}
    \multicolumn{2}{c|}{} &DCB & DTS & A & C & P & S \\
    \hline
    \midrule
   A& Baseline   &            &            & $71.53$         & $75.30$         & $86.71$         & $78.75$         & $78.10$ \\
   B& A w/ DCB    & \checkmark &            & $73.63$         & $75.47$         & 87.43           & $77.81$         & $78.59$   \\
   C& A w/ DTS    &            &\checkmark  & $79.98$         & \bfseries79.31  & \bfseries94.25  & $79.66$         & $83.30$  \\
    
   D& Ours (ME)  & \checkmark & \checkmark & \bfseries81.15  & 78.92           & $93.95$         & \bfseries81.73  & \bfseries83.94 \\

    \bottomrule
    \end{tabular}
    \label{componets}
\end{table}

\begin{figure*}[t!]
    \centering
    
    \subfloat[The loss of `Art painting']{\includegraphics[width=1.8in]{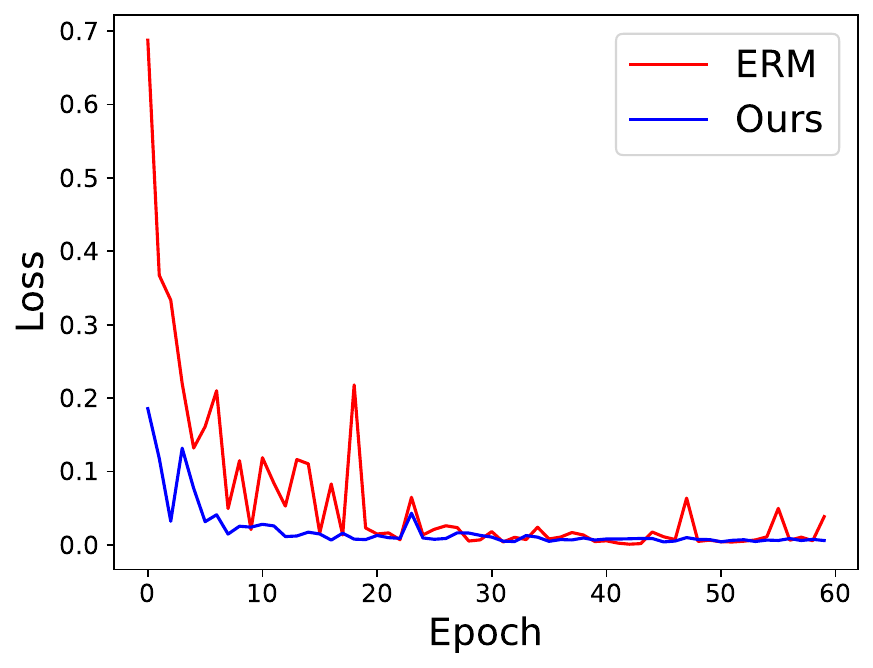}}
        \subfloat[The loss of `Cartoon']{\includegraphics[width=1.8in]{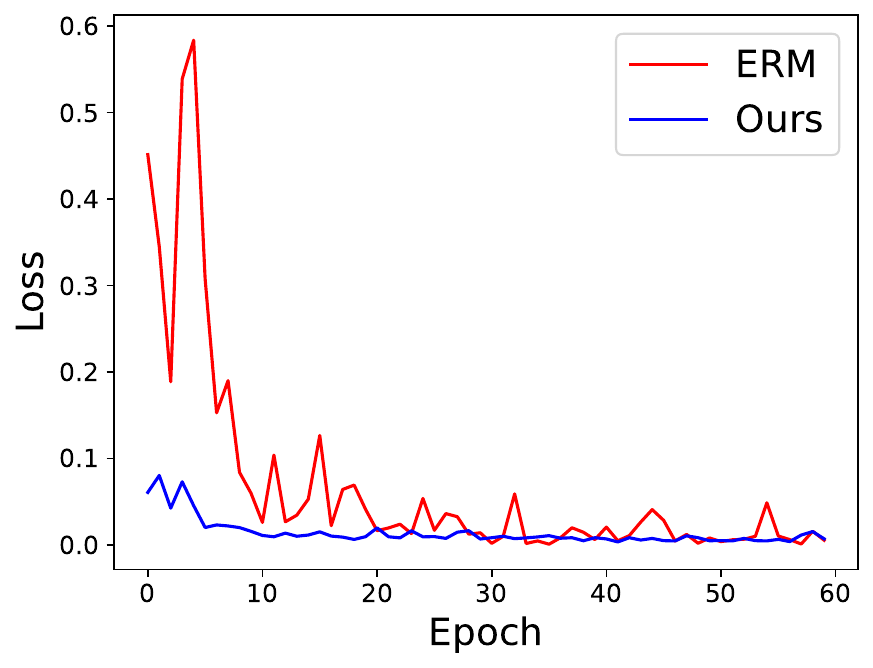}}
        \subfloat[The loss of `Sketch']{\includegraphics[width=1.8in]{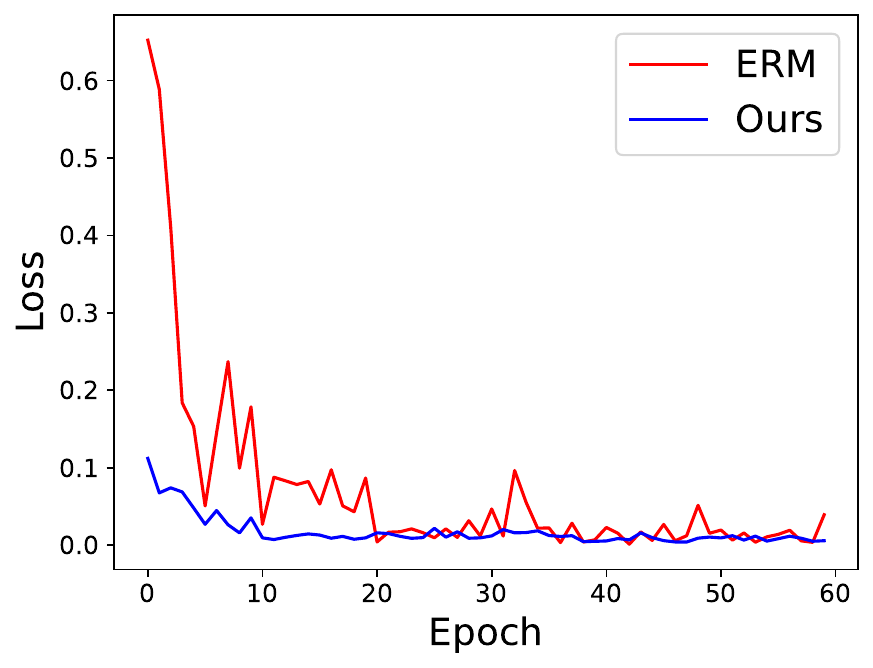}}
        \subfloat[The total loss]{\includegraphics[width=1.8in]{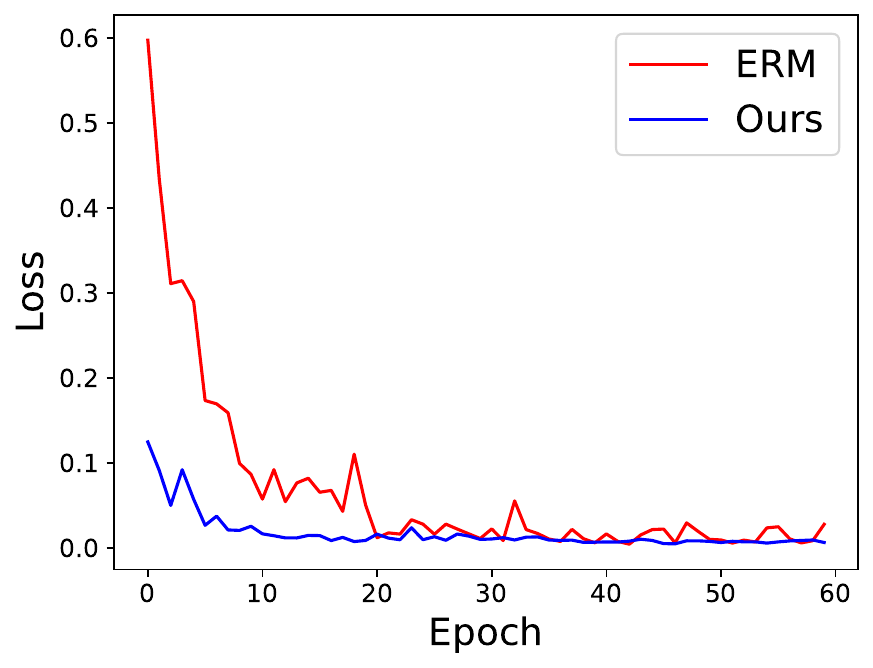}}
    \caption{Loss fluctuation comparison. We plot the training losses of ERM and DTCS (ours) in each source domains on PACS during training, with `Photo' as the target domain. The smoother training loss evolvement verifies the effectiveness of the proposed diverse target supervision for alleviating gradient conflicts.}
    \label{loss compare}
\end{figure*}

\begin{figure*}
    \centering
    \includegraphics[width=0.99\linewidth]{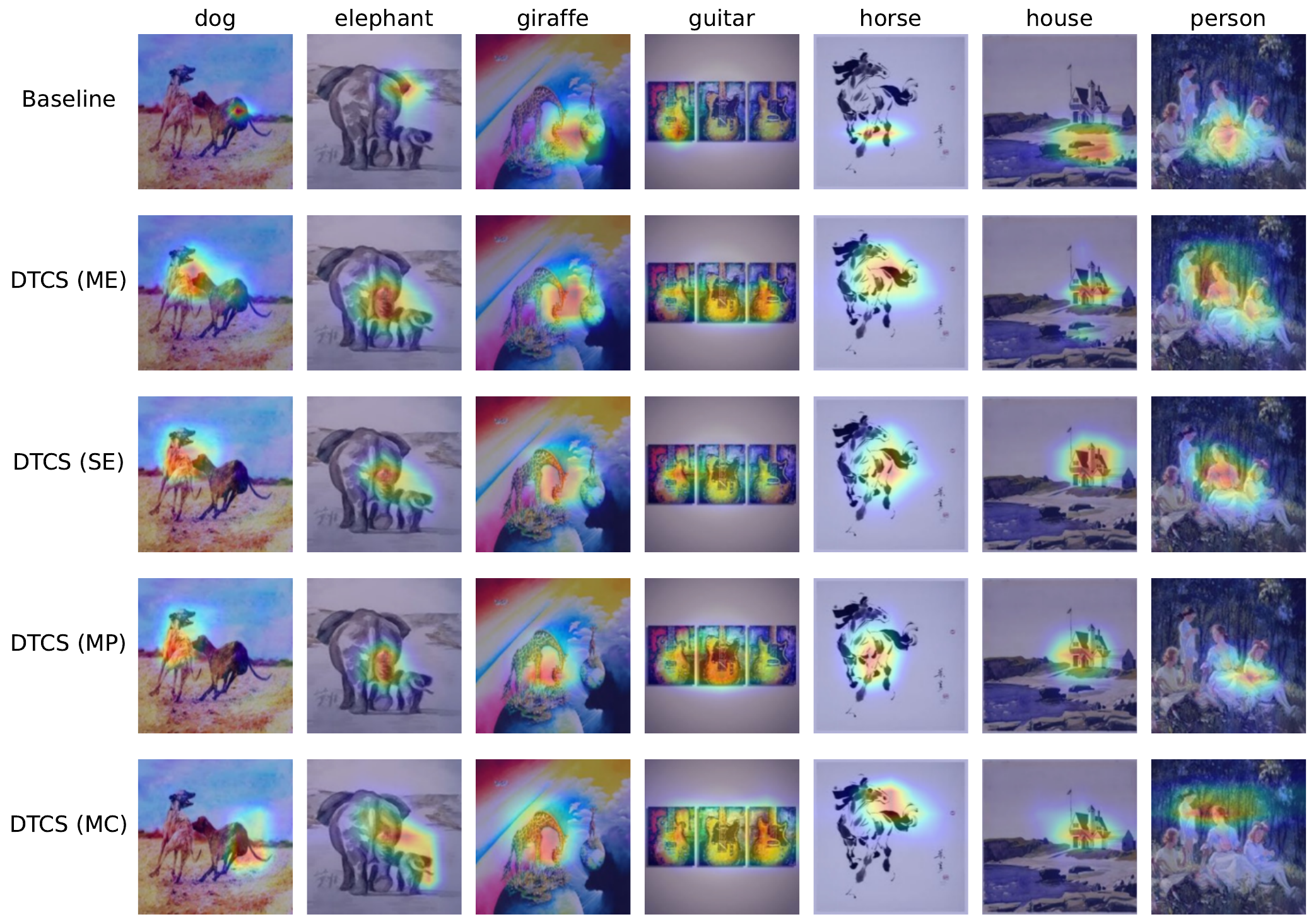}
    \caption{Visualizations of attention maps in the last convolution layer for the baseline and the proposed DTCS on PACS, with `Art painting' as the target domain and images sampled from `Art painting'.}
    \label{heatmap}
\end{figure*}

\noindent\textbf{Diverse Targets Matter.} 
Our proposed DTS shares similarities with knowledge distillation~\cite{hinton2015distilling} in the sense that extra pre-trained models are utilized to generate the soft labels when the diverse target prophets are composed of either multiple or single experts. However, we contend that the effectiveness of our proposed DTS stems from the use of diverse targets rather than the knowledge distillation process of transferring knowledge from a bigger model to a small model. As shown in Section~\ref{exp}, even when the diverse targets are generated by models with comparable or even inferior parameters and  parameter size~(\emph{e.g.}, the hypothesis model of previous epoch or just multiple linear classifiers), we still observe significant improvements in generalization capacities. This suggests that it is the introduction of diverse targets, which helps alleviate the gradient conflicts, that truly matters in our approach.

\begin{table}[t]
    \centering
    \setlength{\tabcolsep}{3pt}
    
    \caption{Comparison of generation performance and stability of our proposed approaches with SOTA methods on PACS. $GS$ measures the generalization stability.}
    \begin{tabular}{l|cccc|c|c}
    \toprule
    \multirow{2}{*}{Model} &  \multicolumn{4}{c|}{Target domain} & \multirow{2}{*}{Avg($\uparrow$)}& \multirow{2}{*}{$GS$($\downarrow$)} \\
    \cmidrule(r){2-5}
    & A & C & P & S &\\
    \hline
    \midrule
    Arg-sum~\cite{mansilla2021domain}        & 75.63         & 77.39       & 92.63  & 72.66        & 79.58      & 8.92\\
   Fish~\cite{shi2022gradient}        & 80.70         & 74.52        & 95.24  & 75.21        & 81.41       & 9.62 \\
    GradCa~\cite{song2023gradca}        & 82.51  & 78.66       & 96.47         & 77.25        & 83.72       & 8.78 \\
    SWAD~\cite{cha2021swad}        & $83.28$         & $74.63$        & \bfseries96.56  & $77.96$        & $83.11$       & 9.65 \\
    PCL~\cite{yao2022pcl}        & \bfseries83.53  & 73.61          & $96.18$         & $77.20$        & $82.63$       & 9.92 \\
    \hline
   {Ours~(ME)}   & 81.15           & \bfseries78.92 & $93.95$         & \bfseries81.73 & \bfseries83.94& \bfseries6.78\\
    {Ours~(SE)}  & 82.28           & 78.11          & $94.61$         & 78.26          & 83.31         & 7.77  \\
    {Ours~(MP)}& 82.28           & 78.24          & $93.53$         & 78.42          & 83.12         & 7.19 \\
    {Ours~(MC)}  & 81.59           & 78.46          & $94.49$         & 79.43          & 83.49         & 7.45 \\
   
    \bottomrule
    \end{tabular}
    \label{single}
\end{table}

\noindent\textbf{Performance on Hard-to-Transfer Domains.} To measure the generalization capacity of models, it is important to consider not only the average of the generalization performance across various domains but also the performance on hard-to-transfer domains which exhibit significant variations in style and shape compared to other domains. As illustrated in TABLE~\ref{single}, our proposed model demonstrates remarkable improvements on these hard-to-transfer domains, namely, `Cartoon' and `Sketch', where existing models do not behave as well as on other domains. This highlights the effectiveness of our model in mitigating the gradient conflicts and preventing the model from excessively focusing on special source domains, thus ensuring the stability of the generalization performance. To quantify the stability of generalization capacity, we introduce the concept of generalization stability~($GS$), defined as:
\begin{equation}
    GS = \sqrt{\sum_{i = 1}^M(GP(i) - \overline{GP})^2},
    \label{stable}
\end{equation}
where $GP(i)$ is the generalization performance on the source domain $i$, $\overline{GP} := \frac{1}{M}\sum_{i =1}^M GP(i)$ is the average of the generalization performances across domains. The generalization stability is also reported in TABLE~\ref{single}. The proposed models with different diverse target prophets consistently exhibit lower $GS$ values compared to state-of-the-art methods, confirming our claim that the proposed model can enhance the performance on hard-to-transfer domains.

\begin{figure}[H]
    \centering
      \vspace{-0.3cm}
    \includegraphics[width=0.7\linewidth]{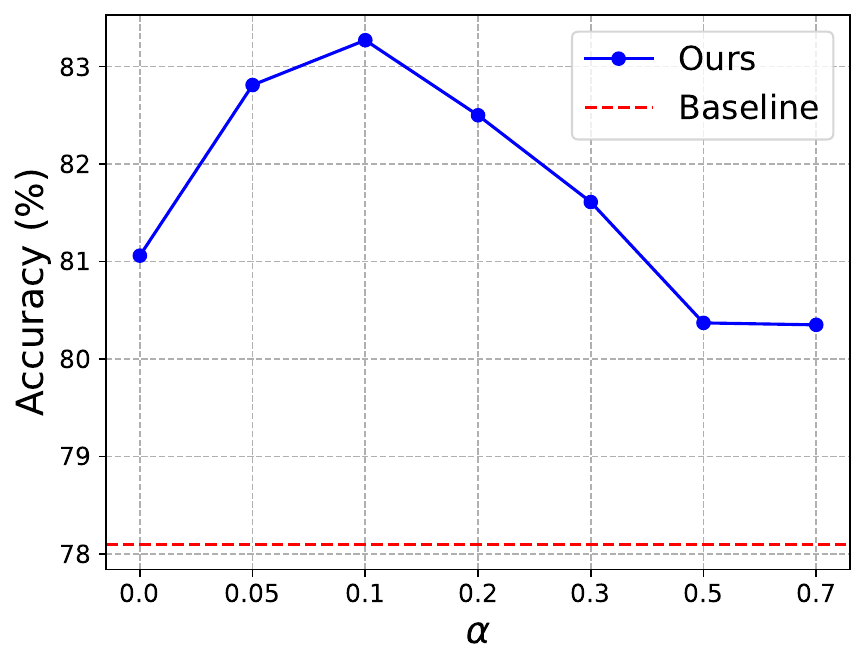}
      \setlength{\abovecaptionskip}{-0.05cm}
    \caption{Generalization performance under different weighting factor $\alpha$.}
    \label{alpha}
\end{figure}

\noindent\textbf{Sensitivity of $\alpha$.}
We check the sensitivity of the trade-off parameter $\alpha$ reflecting the weight of the diverse targets by varying $\alpha \in \{0.0, 0.05, 0.1, 0.2, 0.3, 0.5, 0.7\}$ on PACS. As shown in Fig.~\ref{alpha}, the accuracy of the proposed model first rises and then decreases as $\alpha$ varies, and consistently remains higher than that of the baseline. This finding confirms the necessity of utilizing diverse targets in mitigating gradient conflicts and thereby enhancing generalization capacities.

\noindent\textbf{Sensitivity of $\tau$.} We also conduct experiments on PACS to investigate the influence of the hyper-parameter $\tau$ on the generalization performance. As shown in TABLE~\ref{temper}, DCB consistently improves the generalization performance across different values of $\tau$. Furthermore, an appropriate temperature value can benefit the
performance. Excessively high temperatures, which produce softer probability distributions, may lose significant information for accurate object recognition. Conversely, excessively low temperatures, which produce sharper probability distributions, may not provide sufficient knowledge to effectively mitigate gradient conflicts.
\begin{table}[h]
    \centering
    \setlength{\tabcolsep}{3pt}
    
    \caption{Generalization performance of the proposed model with or without (w/o) diverse contribution balance~(DCB) under various temperatures $\tau$ on PACS.}
    \begin{tabular}{l|c|cccc|c}
    \toprule
    \multirow{2}{*}{Model} & \multirow{2}{*}{$\tau$} & \multicolumn{4}{c|}{Target domain} & \multirow{2}{*}{Avg($\uparrow$)} \\
    \cmidrule(r){3-6}
    & & A & C & P & S \\
    \hline
    \midrule
    DTCS & \multirow{2}{*}{1.0} & $80.52$  & $78.84$ & $93.71$  & $80.43$ & $83.38$ \\
    \quad w/o DCB &              & $80.62$  & $77.52$ & $94.73$  & $79.33$ & $83.05$ \\
    \midrule
    DTCS & \multirow{2}{*}{2.0} & $81.15$  & $78.92$ & $93.95$  & $81.73$ & $83.94$ \\
    \quad w/o DCB &              & $79.98$  & $79.31$ & $94.25$  & $79.66$ & $83.30$ \\
    \midrule
    DTCS & \multirow{2}{*}{3.0} & $80.62$  & $78.71$ & $93.95$  & $81.04$ & $83.58$ \\
    \quad w/o DCB &              & $79.83$  & $77.90$ & $94.25$  & $78.16$ & $82.54$ \\
    \midrule
    DTCS & \multirow{2}{*}{4.5} & $79.44$  & $79.31$ & $93.77$  & $80.55$ & $83.27$ \\
    \quad w/o DCB &              & $78.71$  & $80.20$ & $93.41$  & $80.43$ & $83.19$ \\
    \bottomrule
    \end{tabular}
    \label{temper}
\end{table}

\noindent\textbf{Convergence Analysis.} We analyze the training dynamics of the proposed method on PACS and compare the results with ERM to demonstrate that the proposed strategy can mitigate the gradient conflict caused by the distribution shift. As shown in Fig.~\ref{loss compare}, the training loss of our proposed model exhibits smoother behavior in each source domain compared to ERM, not only during the initial stage but also at convergence. This pattern is also observed for the total loss, demonstrating that the proposed DTS can alleviate the conflict of gradient descent directions. Consequently, the hypothesis model can fully leverage the knowledge across source domains to enhance its generalization capacity in novel environments. To further evaluate the impact of DTS, we compare the standard deviations of losses since hypothesis model converged~(\emph{e.g.,} after 20th epoch in our case, as shown in Fig.\ref{loss compare}) in different source domains, as well as the total loss, with ERM~\cite{vapnik1999overview} and existing gradient manipulation methods~\cite{mansilla2021domain, shi2022gradient} in TABLE~\ref{std}. The significant reductions in standard deviations across all scenarios confirm that the proposed DTS effectively mitigates the gradient conflicts during training.

\begin{table}[]
    \centering
    \caption{Comparison of the loss' standard deviations (Std.) since the hypothesis model converged for ERM, gradient manipulation methods, and our DTCS on PACS, with `Art painting' as the target domain.  DCB means diverse contribution balance.}
    \begin{tabular}{c|c|c|c|c}
    \toprule
    \multirow{2}{*}{Method} & \multicolumn{4}{c}{Std. of loss ($\downarrow$)}\\
    \cmidrule(r){2-5}
    & Cartoon & Photo & Sketch & Total \\
    \midrule
    ERM~\cite{vapnik1999overview}        & 8.71e-3         & 5.34e-3          & 3.47e-2          & 1.32e-2 \\
    Agr-sum~\cite{mansilla2021domain}        & 1.66e-2         & 5.84e-3          & 5.13e-2          & 1.77e-2 \\
    Fish~\cite{shi2022gradient}        & 2.34e-2         & 2.33e-3          & 2.03e-2          & 2.03e-2 \\
    \midrule
    DTCS w/o DCB  &\bfseries3.40e-3 & 2.02e-3          & 4.27e-3          & 2.26e-3\\
    DTCS (ours)  & 3.55e-3         & \bfseries1.97e-3 & \bfseries3.41e-3 & \bfseries2.22e-3\\
    \bottomrule
    
    \end{tabular}
    \label{std}
\end{table}

\noindent\textbf{Difference with Existing Strategies for Gradient Conflicts.}
Existing methods in DG for addressing gradient Conflicts rely on gradient manipulation~\cite{mansilla2021domain, song2023gradca, shi2022gradient}, such as removing or modifying contradictory gradients, or maximizing the inner product of gradients across domains. These methods can be regarded as domain-invariant learning methods, as they are dedicated to seeking gradients that are consistent across domains. However, they encounter the shortcomings associated with domain-invariant learning methods, including indiscriminate removal of domain-specific information, which could lead to increased empirical source risk and reduced within-class diversity. Taking Arg-sum~\cite{mansilla2021domain} for example, as reported in ~\ref{single}, its performance on PACS is unsatisfactory. This can be attributed to the sparse sign-agreement gradients across domains, which severely impacts the gradient updates. As shown in TABLE~\ref{std}, these methods do not exhibit a significant decrease and may even increase the loss' standard deviation compared to ERM, indicating the challenges in simultaneously reducing empirical risks across domains. In contrast, our proposed DTS employs diverse targets to fundamentally solve the gradient conflicts during optimization, significantly mitigating the loss fluctuations. Besides, our proposed model also performs better on hard-to-transfer domains, as demonstrated in TABLE~\ref{single}. We conjecture that this improvement can be attributed to the diverse targets, which divide the total task into non-contradictory sub-tasks, thereby avoiding the disturbance encountered in other gradient manipulation methods~\cite{mansilla2021domain, song2023gradca, shi2022gradient} for learning on hard-to-transfer domains.

\noindent\textbf{Visual Analysis with GradCAM.} To visually verify the superiority of the proposed model, we utilize GradCAM~\cite{selvaraju2017grad} to generate attention maps of the last convolution layer for both the baseline and the proposed methods. As shown in Fig.~\ref{heatmap}, our methods can still capture the category-related features, while the baseline is hindered by optimization risks associated with gradient conflicts. Taking `person' for example, the proposed models with different diverse target prophets capture the physical characteristics of multiple individuals, whereas the baseline primarily focuses on the attire of a single person.

\noindent\textbf{Orthogonal to Existing DG Methods.} The proposed DTS, aiming to alleviate gradient conflicts which is orthogonal to other models in DG that modified the training procedures or architectures, is a versatile strategy and offers many advantages that set it apart from other DG methods. We combine DTS with existing DG methods such as CORAL~\cite{sun2016deep}~(domain invariance method), Mixup~\cite{yan2020improve}~(data augmentation technique), Fish~\cite{shi2022gradient}~(gradient matching strategy), and PCL~\cite{yao2022pcl}~(contrastive learning method) to verify this claim. In this combination, we simply alternate the original cross-entropy loss with our proposed loss in Eq.~\ref{kd_loss} while keeping the weights for their respective objectives unchanged. Furthermore, we do not conduct fine-tuning and just maintain the experiment settings in DTS, including the temperature $\tau$ of 2 and the weighting factor $\alpha$ of 0.1, which further highlights the effectiveness of the proposed module. The results, shown in TABLE~\ref{combine}, consistently demonstrate that DTS brings significant generalization capacity promotion to existing DG techniques, including domain invariance learning, data augmentation, gradient matching, and contrastive learning. Notably, combining CORAL and DTS can incorporate the superiority of promoting domain invariance and mitigating gradient conflicts, exhibiting the best generalization performance. 

\begin{table}[]
    \centering
    \setlength{\tabcolsep}{3pt}
    
    \caption{Combination of the proposed DTS and other methods in DG on PACS. $\dag$: without ensembling technique.}
    \begin{tabular}{l|cccc|c}
    \toprule
    \multirow{2}{*}{Model}  & \multicolumn{4}{c|}{Target domain} & \multirow{2}{*}{Avg($\uparrow$)} \\
    \cmidrule(r){2-5}
    & A & C & P & S \\
    \hline
    \midrule
    CORAL~\cite{sun2016deep}        &  $79.50$  & $76.49$ & $95.58$  & $76.05$ & $81.91$ \\
    \quad w/ DTS  & $82.86$   & $77.19$ & $95.36$  & $78.50$ & $83.48$ \\
    \midrule
    Mixup~\cite{yan2020improve}        &  79.38    & 78.79 & 94.39      & 68.86   & 80.35 \\
    \quad w/ DTS  & 80.84     & 76.51 & 94.79      & 74.71   & 81.71 \\
    \midrule
    Fish~\cite{shi2022gradient}         &  80.70    & 74.52 & 95.24      & 75.21   & 81.41 \\
    \quad w/ DTS  &  81.76    & 76.07 & 94.61      & 76.12   & 82.15 \\
    \midrule
    PCL\dag~\cite{yao2022pcl}     &  83.71    & 72.18 & 94.61      & 71.06   & 80.39 \\
    \quad w/ DTS  & 77.55     & 77.22 & 94.24      & 79.23   & 82.18 \\
    \bottomrule
    \end{tabular}
    \label{combine}
\end{table}

\section{Conclusion}

In this paper, we address the limitations of the commonly used one-hot labels and the equal weights for source domains in DG. Through theoretical analysis, we identify the presence of gradient conflict from the perspective of the bound for empirical source risk, providing a new direction to enhance performance in DG. To promote generalization capacities, we propose a novel approach called Diverse Target and Contribution Scheduling (DTCS), which encompasses two innovative modules: Diverse Target Supervision~(DTS) and Diverse Contribution Balance~(DCB). The proposed DTS leverages diverse targets as guidance to supervise the hypothesis model, and DCB adaptively balances the contributions of source domains according to their relative inverse training rate. In this way, DTCS mitigates the conflict in directions of gradient descent and guarantees the fair contributions of various source domains. Extensive experiments on benchmark datasets demonstrate the effectiveness and advantages of DTCS. Our findings underscore the significance of diverse targets and contributions in improving generalization performance, providing valuable insights for future research in this area.

{
\bibliographystyle{IEEEtran}
\bibliography{dg}

\begin{thebibliography}{10}
\providecommand{\url}[1]{#1}
\csname url@samestyle\endcsname
\providecommand{\newblock}{\relax}
\providecommand{\bibinfo}[2]{#2}
\providecommand{\BIBentrySTDinterwordspacing}{\spaceskip=0pt\relax}
\providecommand{\BIBentryALTinterwordstretchfactor}{4}
\providecommand{\BIBentryALTinterwordspacing}{\spaceskip=\fontdimen2\font plus
\BIBentryALTinterwordstretchfactor\fontdimen3\font minus \fontdimen4\font\relax}
\providecommand{\BIBforeignlanguage}[2]{{%
\expandafter\ifx\csname l@#1\endcsname\relax
\typeout{** WARNING: IEEEtran.bst: No hyphenation pattern has been}%
\typeout{** loaded for the language `#1'. Using the pattern for}%
\typeout{** the default language instead.}%
\else
\language=\csname l@#1\endcsname
\fi
#2}}
\providecommand{\BIBdecl}{\relax}
\BIBdecl

\bibitem{krizhevsky2012imagenet}
A.~Krizhevsky, I.~Sutskever, and G.~E. Hinton, ``Imagenet classification with deep convolutional neural networks,'' \emph{Advances in neural information processing systems}, vol.~25, 2012.

\bibitem{he2016deep}
K.~He, X.~Zhang, S.~Ren, and J.~Sun, ``Deep residual learning for image recognition,'' in \emph{Proceedings of the IEEE conference on computer vision and pattern recognition}, 2016, pp. 770--778.

\bibitem{dosovitskiy2020image}
A.~Dosovitskiy, L.~Beyer, A.~Kolesnikov, D.~Weissenborn, X.~Zhai, T.~Unterthiner, M.~Dehghani, M.~Minderer, G.~Heigold, S.~Gelly \emph{et~al.}, ``An image is worth 16x16 words: Transformers for image recognition at scale,'' in \emph{International Conference on Learning Representations}, 2020.

\bibitem{deng2022dynamic}
Z.~Deng, K.~Zhou, D.~Li, J.~He, Y.-Z. Song, and T.~Xiang, ``Dynamic instance domain adaptation,'' \emph{IEEE Transactions on Image Processing}, vol.~31, pp. 4585--4597, 2022.

\bibitem{feng2022dmt}
Z.~Feng, Q.~Zhou, Q.~Gu, X.~Tan, G.~Cheng, X.~Lu, J.~Shi, and L.~Ma, ``Dmt: Dynamic mutual training for semi-supervised learning,'' \emph{Pattern Recognition}, p. 108777, 2022.

\bibitem{song2023rethinking}
Y.~Song, Q.~Zhou, and L.~Ma, ``Rethinking implicit neural representations for vision learners,'' in \emph{ICASSP 2023-2023 IEEE International Conference on Acoustics, Speech and Signal Processing (ICASSP)}.\hskip 1em plus 0.5em minus 0.4em\relax IEEE, 2023, pp. 1--5.

\bibitem{ren2015faster}
S.~Ren, K.~He, R.~Girshick, and J.~Sun, ``Faster r-cnn: Towards real-time object detection with region proposal networks,'' \emph{Advances in neural information processing systems}, vol.~28, 2015.

\bibitem{lin2017feature}
T.-Y. Lin, P.~Doll{\'a}r, R.~Girshick, K.~He, B.~Hariharan, and S.~Belongie, ``Feature pyramid networks for object detection,'' in \emph{Proceedings of the IEEE conference on computer vision and pattern recognition}, 2017, pp. 2117--2125.

\bibitem{redmon2016you}
J.~Redmon, S.~Divvala, R.~Girshick, and A.~Farhadi, ``You only look once: Unified, real-time object detection,'' in \emph{Proceedings of the IEEE conference on computer vision and pattern recognition}, 2016, pp. 779--788.

\bibitem{lim2023ernet}
J.~Lim, V.~M. Baskaran, J.~M.-Y. Lim, K.~Wong, J.~See, and M.~Tistarelli, ``Ernet: An efficient and reliable human-object interaction detection network,'' \emph{IEEE Transactions on Image Processing}, vol.~32, pp. 964--979, 2023.

\bibitem{yang2022efficient}
Z.~Yang, C.~Zhang, R.~Li, Y.~Xu, and G.~Lin, ``Efficient few-shot object detection via knowledge inheritance,'' \emph{IEEE Transactions on Image Processing}, vol.~32, pp. 321--334, 2022.

\bibitem{he2021end}
L.~He, Q.~Zhou, X.~Li, L.~Niu, G.~Cheng, X.~Li, W.~Liu, Y.~Tong, L.~Ma, and L.~Zhang, ``End-to-end video object detection with spatial-temporal transformers,'' in \emph{Proceedings of the 29th ACM International Conference on Multimedia (ACM MM)}, 2021, pp. 1507--1516.

\bibitem{zhou2022transvod}
Q.~Zhou, X.~Li, L.~He, Y.~Yang, G.~Cheng, Y.~Tong, L.~Ma, and D.~Tao, ``Transvod: end-to-end video object detection with spatial-temporal transformers,'' \emph{IEEE Transactions on Pattern Analysis and Machine Intelligence}, vol.~45, no.~6, pp. 7853--7869, 2023.

\bibitem{xu2021semi}
H.~Xu, F.~Liu, Q.~Zhou, J.~Hao, Z.~Cao, Z.~Feng, and L.~Ma, ``Semi-supervised 3d object detection via adaptive pseudo-labeling,'' in \emph{2021 IEEE International Conference on Image Processing (ICIP)}.\hskip 1em plus 0.5em minus 0.4em\relax IEEE, 2021, pp. 3183--3187.

\bibitem{long2015fully}
J.~Long, E.~Shelhamer, and T.~Darrell, ``Fully convolutional networks for semantic segmentation,'' in \emph{Proceedings of the IEEE conference on computer vision and pattern recognition}, 2015, pp. 3431--3440.

\bibitem{ronneberger2015u}
O.~Ronneberger, P.~Fischer, and T.~Brox, ``U-net: Convolutional networks for biomedical image segmentation,'' in \emph{Medical Image Computing and Computer-Assisted Intervention--MICCAI 2015: 18th International Conference, Munich, Germany, October 5-9, 2015, Proceedings, Part III 18}.\hskip 1em plus 0.5em minus 0.4em\relax Springer, 2015, pp. 234--241.

\bibitem{chen2018encoder}
L.-C. Chen, Y.~Zhu, G.~Papandreou, F.~Schroff, and H.~Adam, ``Encoder-decoder with atrous separable convolution for semantic image segmentation,'' in \emph{Proceedings of the European conference on computer vision (ECCV)}, 2018, pp. 801--818.

\bibitem{ramesh2022hierarchical}
A.~Ramesh, P.~Dhariwal, A.~Nichol, C.~Chu, and M.~Chen, ``Hierarchical text-conditional image generation with clip latents,'' \emph{arXiv preprint arXiv:2204.06125}, 2022.

\bibitem{yuan2022birds}
B.~Yuan, D.~Zhao, S.~Shao, Z.~Yuan, and C.~Wang, ``Birds of a feather flock together: Category-divergence guidance for domain adaptive segmentation,'' \emph{IEEE Transactions on Image Processing}, vol.~31, pp. 2878--2892, 2022.

\bibitem{beery2018recognition}
S.~Beery, G.~Van~Horn, and P.~Perona, ``Recognition in terra incognita,'' in \emph{Proceedings of the European conference on computer vision (ECCV)}, 2018, pp. 456--473.

\bibitem{li2020domain}
H.~Li, Y.~Wang, R.~Wan, S.~Wang, T.-Q. Li, and A.~Kot, ``Domain generalization for medical imaging classification with linear-dependency regularization,'' \emph{Advances in Neural Information Processing Systems}, vol.~33, pp. 3118--3129, 2020.

\bibitem{ding2022word}
L.~Ding, D.~Yu, J.~Xie, W.~Guo, S.~Hu, M.~Liu, L.~Kong, H.~Dai, Y.~Bao, and B.~Jiang, ``Word embeddings via causal inference: Gender bias reducing and semantic information preserving,'' in \emph{Proceedings of the AAAI Conference on Artificial Intelligence}, vol.~36, no.~11, 2022, pp. 11\,864--11\,872.

\bibitem{zhao2022style}
Y.~Zhao, Z.~Zhong, N.~Zhao, N.~Sebe, and G.~H. Lee, ``Style-hallucinated dual consistency learning for domain generalized semantic segmentation,'' in \emph{European Conference on Computer Vision}.\hskip 1em plus 0.5em minus 0.4em\relax Springer, 2022, pp. 535--552.

\bibitem{geirhos2020shortcut}
R.~Geirhos, J.-H. Jacobsen, C.~Michaelis, R.~Zemel, W.~Brendel, M.~Bethge, and F.~A. Wichmann, ``Shortcut learning in deep neural networks,'' \emph{Nature Machine Intelligence}, vol.~2, no.~11, pp. 665--673, 2020.

\bibitem{zhang2021can}
D.~Zhang, K.~Ahuja, Y.~Xu, Y.~Wang, and A.~Courville, ``Can subnetwork structure be the key to out-of-distribution generalization?'' in \emph{International Conference on Machine Learning}.\hskip 1em plus 0.5em minus 0.4em\relax PMLR, 2021, pp. 12\,356--12\,367.

\bibitem{kong2022partial}
L.~Kong, S.~Xie, W.~Yao, Y.~Zheng, G.~Chen, P.~Stojanov, V.~Akinwande, and K.~Zhang, ``Partial disentanglement for domain adaptation,'' in \emph{International Conference on Machine Learning}.\hskip 1em plus 0.5em minus 0.4em\relax PMLR, 2022, pp. 11\,455--11\,472.

\bibitem{stojanov2021domain}
P.~Stojanov, Z.~Li, M.~Gong, R.~Cai, J.~Carbonell, and K.~Zhang, ``Domain adaptation with invariant representation learning: What transformations to learn?'' \emph{Advances in Neural Information Processing Systems}, vol.~34, pp. 24\,791--24\,803, 2021.

\bibitem{wang2022cluster}
S.~Wang, D.~Zhao, C.~Zhang, Y.~Guo, Q.~Zang, Y.~Gu, Y.~Li, and L.~Jiao, ``Cluster alignment with target knowledge mining for unsupervised domain adaptation semantic segmentation,'' \emph{IEEE Transactions on Image Processing}, vol.~31, pp. 7403--7418, 2022.

\bibitem{gu2021pit}
Q.~Gu, Q.~Zhou, M.~Xu, Z.~Feng, G.~Cheng, X.~Lu, J.~Shi, and L.~Ma, ``Pit: Position-invariant transform for cross-fov domain adaptation,'' in \emph{Proceedings of the IEEE/CVF International Conference on Computer Vision (ICCV)}, 2021, pp. 8761--8770.

\bibitem{zhou2022context}
Q.~Zhou, Z.~Feng, Q.~Gu, J.~Pang, G.~Cheng, X.~Lu, J.~Shi, and L.~Ma, ``Context-aware mixup for domain adaptive semantic segmentation,'' \emph{IEEE Transactions on Circuits and Systems for Video Technology}, vol.~33, no.~2, pp. 804--817, 2022.

\bibitem{zhou2022generative}
Q.~Zhou, K.-Y. Zhang, T.~Yao, R.~Yi, K.~Sheng, S.~Ding, and L.~Ma, ``Generative domain adaptation for face anti-spoofing,'' in \emph{European Conference on Computer Vision (ECCV)}.\hskip 1em plus 0.5em minus 0.4em\relax Springer, 2022, pp. 335--356.

\bibitem{zhou2022uncertainty}
Q.~Zhou, Z.~Feng, Q.~Gu, G.~Cheng, X.~Lu, J.~Shi, and L.~Ma, ``Uncertainty-aware consistency regularization for cross-domain semantic segmentation,'' \emph{Computer Vision and Image Understanding}, vol. 221, p. 103448, 2022.

\bibitem{zhou2022domain}
Q.~Zhou, C.~Zhuang, R.~Yi, X.~Lu, and L.~Ma, ``Domain adaptive semantic segmentation via regional contrastive consistency regularization,'' in \emph{2022 IEEE International Conference on Multimedia and Expo (ICME)}.\hskip 1em plus 0.5em minus 0.4em\relax IEEE, 2022, pp. 01--06.

\bibitem{zhou2023self}
Q.~Zhou, Q.~Gu, J.~Pang, X.~Lu, and L.~Ma, ``Self-adversarial disentangling for specific domain adaptation,'' \emph{IEEE Transactions on Pattern Analysis and Machine Intelligence}, vol.~45, no.~7, pp. 8954--8968, 2023.

\bibitem{gulrajani2020search}
I.~Gulrajani and D.~Lopez-Paz, ``In search of lost domain generalization,'' in \emph{International Conference on Learning Representations}, 2020.

\bibitem{wang2022generalizing}
J.~Wang, C.~Lan, C.~Liu, Y.~Ouyang, T.~Qin, W.~Lu, Y.~Chen, W.~Zeng, and P.~Yu, ``Generalizing to unseen domains: A survey on domain generalization,'' \emph{IEEE Transactions on Knowledge and Data Engineering}, 2022.

\bibitem{wang2022contrastive}
Y.~Wang, F.~Liu, Z.~Chen, Y.-C. Wu, J.~Hao, G.~Chen, and P.-A. Heng, ``Contrastive-ace: Domain generalization through alignment of causal mechanisms,'' \emph{IEEE Transactions on Image Processing}, vol.~32, pp. 235--250, 2022.

\bibitem{xia2023generative}
H.~Xia, T.~Jing, and Z.~Ding, ``Generative inference network for imbalanced domain generalization,'' \emph{IEEE Transactions on Image Processing}, vol.~32, pp. 1694--1704, 2023.

\bibitem{zhou2023instance}
Q.~Zhou, K.-Y. Zhang, T.~Yao, X.~Lu, R.~Yi, S.~Ding, and L.~Ma, ``Instance-aware domain generalization for face anti-spoofing,'' in \emph{Proceedings of the IEEE/CVF Conference on Computer Vision and Pattern Recognition (CVPR)}, 2023, pp. 20\,453--20\,463.

\bibitem{zhou2022adaptive}
Q.~Zhou, K.-Y. Zhang, T.~Yao, R.~Yi, S.~Ding, and L.~Ma, ``Adaptive mixture of experts learning for generalizable face anti-spoofing,'' in \emph{Proceedings of the 30th ACM International Conference on Multimedia}, 2022, pp. 6009--6018.

\bibitem{ganin2016domain}
Y.~Ganin, E.~Ustinova, H.~Ajakan, P.~Germain, H.~Larochelle, F.~Laviolette, M.~Marchand, and V.~Lempitsky, ``Domain-adversarial training of neural networks,'' \emph{The journal of machine learning research}, vol.~17, no.~1, pp. 2096--2030, 2016.

\bibitem{li2018deep}
Y.~Li, X.~Tian, M.~Gong, Y.~Liu, T.~Liu, K.~Zhang, and D.~Tao, ``Deep domain generalization via conditional invariant adversarial networks,'' in \emph{Proceedings of the European Conference on Computer Vision (ECCV)}, 2018, pp. 624--639.

\bibitem{zhao2020domain}
S.~Zhao, M.~Gong, T.~Liu, H.~Fu, and D.~Tao, ``Domain generalization via entropy regularization,'' \emph{Advances in Neural Information Processing Systems}, vol.~33, pp. 16\,096--16\,107, 2020.

\bibitem{yao2022pcl}
X.~Yao, Y.~Bai, X.~Zhang, Y.~Zhang, Q.~Sun, R.~Chen, R.~Li, and B.~Yu, ``Pcl: Proxy-based contrastive learning for domain generalization,'' in \emph{Proceedings of the IEEE/CVF Conference on Computer Vision and Pattern Recognition}, 2022, pp. 7097--7107.

\bibitem{mahajan2021domain}
D.~Mahajan, S.~Tople, and A.~Sharma, ``Domain generalization using causal matching,'' in \emph{International Conference on Machine Learning}.\hskip 1em plus 0.5em minus 0.4em\relax PMLR, 2021, pp. 7313--7324.

\bibitem{vapnik1999overview}
V.~N. Vapnik, ``An overview of statistical learning theory,'' \emph{IEEE transactions on neural networks}, vol.~10, no.~5, pp. 988--999, 1999.

\bibitem{li2018domain}
H.~Li, S.~J. Pan, S.~Wang, and A.~C. Kot, ``Domain generalization with adversarial feature learning,'' in \emph{Proceedings of the IEEE conference on computer vision and pattern recognition}, 2018, pp. 5400--5409.

\bibitem{yan2020improve}
S.~Yan, H.~Song, N.~Li, L.~Zou, and L.~Ren, ``Improve unsupervised domain adaptation with mixup training,'' \emph{arXiv preprint arXiv:2001.00677}, 2020.

\bibitem{meng2022attention}
R.~Meng, X.~Li, W.~Chen, S.~Yang, J.~Song, X.~Wang, L.~Zhang, M.~Song, D.~Xie, and S.~Pu, ``Attention diversification for domain generalization,'' in \emph{European conference on computer vision}.\hskip 1em plus 0.5em minus 0.4em\relax Springer, 2022, pp. 322--340.

\bibitem{li2022sparse}
B.~Li, Y.~Shen, J.~Yang, Y.~Wang, J.~Ren, T.~Che, J.~Zhang, and Z.~Liu, ``Sparse mixture-of-experts are domain generalizable learners,'' in \emph{International Conference on Learning Representations}, 2023.

\bibitem{li2022invariant}
B.~Li, Y.~Shen, Y.~Wang, W.~Zhu, D.~Li, K.~Keutzer, and H.~Zhao, ``Invariant information bottleneck for domain generalization,'' in \emph{Proceedings of the AAAI Conference on Artificial Intelligence}, vol.~36, no.~7, 2022, pp. 7399--7407.

\bibitem{deng2009imagenet}
J.~Deng, W.~Dong, R.~Socher, L.-J. Li, K.~Li, and L.~Fei-Fei, ``Imagenet: A large-scale hierarchical image database,'' in \emph{2009 IEEE conference on computer vision and pattern recognition}.\hskip 1em plus 0.5em minus 0.4em\relax Ieee, 2009, pp. 248--255.

\bibitem{zhang2022towards}
X.~Zhang, L.~Zhou, R.~Xu, P.~Cui, Z.~Shen, and H.~Liu, ``Towards unsupervised domain generalization,'' in \emph{Proceedings of the IEEE/CVF Conference on Computer Vision and Pattern Recognition}, 2022, pp. 4910--4920.

\bibitem{gan2022decorate}
Y.~Gan, X.~Ma, Y.~Lou, Y.~Bai, R.~Zhang, N.~Shi, and L.~Luo, ``Decorate the newcomers: Visual domain prompt for continual test time adaptation,'' \emph{arXiv preprint arXiv:2212.04145}, 2022.

\bibitem{volpi2018generalizing}
R.~Volpi, H.~Namkoong, O.~Sener, J.~C. Duchi, V.~Murino, and S.~Savarese, ``Generalizing to unseen domains via adversarial data augmentation,'' \emph{Advances in neural information processing systems}, vol.~31, 2018.

\bibitem{shankar2018generalizing}
S.~Shankar, V.~Piratla, S.~Chakrabarti, S.~Chaudhuri, P.~Jyothi, and S.~Sarawagi, ``Generalizing across domains via cross-gradient training,'' in \emph{International Conference on Learning Representations}, 2018.

\bibitem{zhou2020domain}
K.~Zhou, Y.~Yang, Y.~Qiao, and T.~Xiang, ``Domain generalization with mixstyle,'' in \emph{International Conference on Learning Representations}, 2021.

\bibitem{anoosheh2018combogan}
A.~Anoosheh, E.~Agustsson, R.~Timofte, and L.~Van~Gool, ``Combogan: Unrestrained scalability for image domain translation,'' in \emph{Proceedings of the IEEE conference on computer vision and pattern recognition workshops}, 2018, pp. 783--790.

\bibitem{zhou2020deep}
K.~Zhou, Y.~Yang, T.~Hospedales, and T.~Xiang, ``Deep domain-adversarial image generation for domain generalisation,'' in \emph{Proceedings of the AAAI Conference on Artificial Intelligence}, vol.~34, no.~07, 2020, pp. 13\,025--13\,032.

\bibitem{kim2021selfreg}
D.~Kim, Y.~Yoo, S.~Park, J.~Kim, and J.~Lee, ``Selfreg: Self-supervised contrastive regularization for domain generalization,'' in \emph{Proceedings of the IEEE/CVF International Conference on Computer Vision}, 2021, pp. 9619--9628.

\bibitem{cha2021swad}
J.~Cha, S.~Chun, K.~Lee, H.-C. Cho, S.~Park, Y.~Lee, and S.~Park, ``Swad: Domain generalization by seeking flat minima,'' \emph{Advances in Neural Information Processing Systems}, vol.~34, pp. 22\,405--22\,418, 2021.

\bibitem{zhou2021domain}
K.~Zhou, Y.~Yang, Y.~Qiao, and T.~Xiang, ``Domain adaptive ensemble learning,'' \emph{IEEE Transactions on Image Processing}, vol.~30, pp. 8008--8018, 2021.

\bibitem{mansilla2021domain}
L.~Mansilla, R.~Echeveste, D.~H. Milone, and E.~Ferrante, ``Domain generalization via gradient surgery,'' in \emph{Proceedings of the IEEE/CVF International Conference on Computer Vision}, 2021, pp. 6630--6638.

\bibitem{song2023gradca}
Y.~Song, Z.~Liu, R.~Tang, G.~Duan, and J.~Tan, ``Gradca: Generalizing to unseen domains via gradient calibration,'' \emph{Neurocomputing}, 2023.

\bibitem{shi2022gradient}
Y.~Shi, J.~Seely, P.~Torr, S.~N, A.~Hannun, N.~Usunier, and G.~Synnaeve, ``Gradient matching for domain generalization,'' in \emph{International Conference on Learning Representations}, 2022.

\bibitem{muller1997integral}
A.~M{\"u}ller, ``Integral probability metrics and their generating classes of functions,'' \emph{Advances in Applied Probability}, vol.~29, no.~2, pp. 429--443, 1997.

\bibitem{ben2010theory}
S.~Ben{-}David, J.~Blitzer, K.~Crammer, A.~Kulesza, F.~Pereira, and J.~W. Vaughan, ``A theory of learning from different domains,'' \emph{Mach. Learn.}, vol.~79, no. 1-2, pp. 151--175, 2010.

\bibitem{chen2018gradnorm}
Z.~Chen, V.~Badrinarayanan, C.-Y. Lee, and A.~Rabinovich, ``Gradnorm: Gradient normalization for adaptive loss balancing in deep multitask networks,'' in \emph{International conference on machine learning}.\hskip 1em plus 0.5em minus 0.4em\relax PMLR, 2018, pp. 794--803.

\bibitem{liu2019end}
S.~Liu, E.~Johns, and A.~J. Davison, ``End-to-end multi-task learning with attention,'' in \emph{Proceedings of the IEEE/CVF conference on computer vision and pattern recognition}, 2019, pp. 1871--1880.

\bibitem{arjovsky2019invariant}
M.~Arjovsky, L.~Bottou, I.~Gulrajani, and D.~Lopez-Paz, ``Invariant risk minimization,'' \emph{arXiv preprint arXiv:1907.02893}, 2019.

\bibitem{sagawa2019distributionally}
S.~Sagawa, P.~W. Koh, T.~B. Hashimoto, and P.~Liang, ``Distributionally robust neural networks,'' in \emph{International Conference on Learning Representations}, 2019.

\bibitem{krueger2021out}
D.~Krueger, E.~Caballero, J.-H. Jacobsen, A.~Zhang, J.~Binas, D.~Zhang, R.~Le~Priol, and A.~Courville, ``Out-of-distribution generalization via risk extrapolation (rex),'' in \emph{International Conference on Machine Learning}.\hskip 1em plus 0.5em minus 0.4em\relax PMLR, 2021, pp. 5815--5826.

\bibitem{huang2020self}
Z.~Huang, H.~Wang, E.~P. Xing, and D.~Huang, ``Self-challenging improves cross-domain generalization,'' in \emph{Computer Vision--ECCV 2020: 16th European Conference, Glasgow, UK, August 23--28, 2020, Proceedings, Part II}, 2020, pp. 124--140.

\bibitem{blanchard2021domain}
G.~Blanchard, A.~A. Deshmukh, {\"U}.~Dogan, G.~Lee, and C.~Scott, ``Domain generalization by marginal transfer learning,'' \emph{The Journal of Machine Learning Research}, vol.~22, no.~1, pp. 46--100, 2021.

\bibitem{nam2021reducing}
H.~Nam, H.~Lee, J.~Park, W.~Yoon, and D.~Yoo, ``Reducing domain gap by reducing style bias,'' in \emph{Proceedings of the IEEE/CVF Conference on Computer Vision and Pattern Recognition}, 2021, pp. 8690--8699.

\bibitem{zhang2021adaptive}
M.~Zhang, H.~Marklund, N.~Dhawan, A.~Gupta, S.~Levine, and C.~Finn, ``Adaptive risk minimization: Learning to adapt to domain shift,'' \emph{Advances in Neural Information Processing Systems}, vol.~34, pp. 23\,664--23\,678, 2021.

\bibitem{li2017deeper}
D.~Li, Y.~Yang, Y.-Z. Song, and T.~M. Hospedales, ``Deeper, broader and artier domain generalization,'' in \emph{Proceedings of the IEEE international conference on computer vision}, 2017, pp. 5542--5550.

\bibitem{venkateswara2017deep}
H.~Venkateswara, J.~Eusebio, S.~Chakraborty, and S.~Panchanathan, ``Deep hashing network for unsupervised domain adaptation,'' in \emph{Proceedings of the IEEE conference on computer vision and pattern recognition}, 2017, pp. 5018--5027.

\bibitem{fang2013unbiased}
C.~Fang, Y.~Xu, and D.~N. Rockmore, ``Unbiased metric learning: On the utilization of multiple datasets and web images for softening bias,'' in \emph{Proceedings of the IEEE International Conference on Computer Vision}, 2013, pp. 1657--1664.

\bibitem{hinton2015distilling}
G.~Hinton, O.~Vinyals, and J.~Dean, ``Distilling the knowledge in a neural network,'' \emph{arXiv preprint arXiv:1503.02531}, 2015.

\bibitem{selvaraju2017grad}
R.~R. Selvaraju, M.~Cogswell, A.~Das, R.~Vedantam, D.~Parikh, and D.~Batra, ``Grad-cam: Visual explanations from deep networks via gradient-based localization,'' in \emph{Proceedings of the IEEE international conference on computer vision}, 2017, pp. 618--626.

\bibitem{sun2016deep}
B.~Sun and K.~Saenko, ``Deep coral: Correlation alignment for deep domain adaptation,'' in \emph{Computer Vision--ECCV 2016 Workshops: Amsterdam, The Netherlands, October 8-10 and 15-16, 2016, Proceedings, Part III 14}.\hskip 1em plus 0.5em minus 0.4em\relax Springer, 2016, pp. 443--450.

\bibitem{panaretos2019statistical}
V.~M. Panaretos and Y.~Zemel, ``Statistical aspects of wasserstein distances,'' \emph{Annual review of statistics and its application}, vol.~6, pp. 405--431, 2019.

\bibitem{tolstikhin2016minimax}
I.~O. Tolstikhin, B.~K. Sriperumbudur, and B.~Sch{\"o}lkopf, ``Minimax estimation of maximum mean discrepancy with radial kernels,'' \emph{Advances in Neural Information Processing Systems}, vol.~29, 2016.

\bibitem{vapnik1971uniform}
V.~Vapnik and A.~Y. Chervonenkis, ``On the uniform convergence of relative frequencies of events to their probabilities,'' \emph{Theory of Probability and its Applications}, vol.~16, no.~2, p. 264, 1971.

\end{thebibliography}
}

\newpage
\appendix
\section{Proofs}

\subsection{Definitions and lemmas}
\begin{definition}
Let ${\rm G}$ be a function family including functions $g: X \rightarrow \mathbb{R}$. For two distributions $P$, $Q$ over $X$, the Integral Probability Metric can be defined as:
\begin{equation}
    {\rm IPM_G}(P, Q) = \sup_{g \in {\rm G}}|\int_X g(x)(P(x) - Q(x))\mathrm{d}x|.
\end{equation}
\end{definition}

${\rm IPM_G}(\cdot,\cdot)$ defines a measure of discrepancy for distributions on the probability function over $X$. With sufficiently large function family ${\rm G}$, ${\rm IPM_G}(\cdot,\cdot)$ can properly evaluate the distribution gap between two distributions~\cite{muller1997integral}. Given different function families, ${\rm IPM_G}(\cdot,\cdot)$ can derive various effective metric for distribution gap, like Wasserstein Distance~\cite{panaretos2019statistical} and Maximum Mean Discrepancy~\cite{tolstikhin2016minimax}.

\begin{lemma}
Consider a set of $M$ mini-batches $\{b_i\}_{i = 1}^{M}$ sampled from $M$ source domains such that the batch for training $B := \{b_i|i = 1, 2, \cdots, M\}$, $n = |b_1|= |b_2| = \cdots =|b_M|$. Let $\mathcal{H}$ be a hypothesis space of VC dimension d. If $\hat{h} \in \mathcal{H}$ is the empirical minimizer of $\frac{1}{M}\sum_{i = 1}^M \hat{\mathcal{E}}_{S_i}$ over the $M$ mini-batches. Then for any $\delta \in (0, 1)$, the following bound holds with probability at least $1-\delta$,
\begin{equation}
    \begin{aligned}
        |\frac{1}{M}\sum_{i = 1}^M \mathcal{E}_{S_i} - \frac{1}{M}\sum_{i = 1}^M \hat{\mathcal{E}}_{S_i}| \leq
        \sqrt{\frac{8d\ln{\frac{2eMn}{d}} + 8\ln{\frac{4}{\delta}}}{Mn}}.
    \end{aligned}
\end{equation}
\label{VC}
\end{lemma}
Lemma~\ref{VC} directly comes from the Vapnik-Chervonenkis (VC) Inequality~\cite{vapnik1971uniform}, bounding the difference between risk and empirical risk by the number of samples and the Vapnik-Chervonenkis dimension. We provide the simple proof here:
\begin{proof}
    \begin{equation}
        \begin{aligned}
           &P(|\frac{1}{M}\sum_{i = 1}^M \mathcal{E}_{S_i} - \frac{1}{M}\sum_{i = 1}^M \hat{\mathcal{E}}_{S_i}| > \epsilon)\\
           &\leq 4m_{\mathcal{H}}(2Mn)\exp({-\frac{Mn\epsilon^2}{8}})\\
           &\leq 4(\frac{2eMn}{d})^d\exp({-\frac{Mn\epsilon^2}{8}}),
        \end{aligned}
    \end{equation}
where $m_{\mathcal{H}}(2Mn)$ is the growth function for the hypothesis space $\mathcal{H}$ with $2Mn$ samples. Based on this fact, we can derive the following inequality:
\begin{equation}
        \begin{aligned}
           &P(|\frac{1}{M}\sum_{i = 1}^M \mathcal{E}_{S_i} - \frac{1}{M}\sum_{i = 1}^M \hat{\mathcal{E}}_{S_i}| \leq \epsilon)\\
           &\geq 1 - 4(\frac{2eMn}{d})^d\exp({-\frac{Mn\epsilon^2}{8}}),
        \end{aligned}
        \label{pro}
    \end{equation}
then we introduce $\delta \in (0, 1)$ and set $4(\frac{2eMn}{d})^d\exp({-\frac{Mn\epsilon^2}{8}}) = \delta$, yielding the result
\begin{equation}
        \begin{aligned}
           \epsilon = \sqrt{\frac{8d\ln{\frac{2eMn}{d}} + 8\ln{\frac{4}{\delta}}}{Mn}}.
        \end{aligned}
        \label{value}
    \end{equation}
Substituting Eq.~\ref{value} into Eq.~\ref{pro} concludes the proof.
\end{proof}

\subsection{Proof of Theorem 1}\label{appe}
\begin{proof}
    Let $g(x):=\ell (\hat{h}(x), y) \in {\rm G}$. Then we have:
    \begin{equation}
        \begin{aligned}
        & \mathcal{E}_{S_j}(\hat{h}) \\
         =&\mathcal{E}_{S_j}(\hat{h})- \sum_{i \neq j}\frac{1}{M - 1}\mathcal{E}_{S_i}(\hat{h})
        + \sum_{i \neq j}\frac{1}{M - 1}\mathcal{E}_{S_i}(\hat{h}) \\
           \leq &|\sum_{i \neq j}\frac{1}{M - 1}\mathcal{E}_{S_i}(\hat{h}) - \mathcal{E}_{S_j}(\hat{h})| + \sum_{i \neq j}\frac{1}{M - 1}\mathcal{E}_{S_i}(\hat{h}) \\
            \leq &\sum_{i \neq j}\frac{1}{M - 1}|\mathcal{E}_{S_i}(\hat{h}) - \mathcal{E}_{S_j}(\hat{h})| + \sum_{i \neq j}\frac{1}{M - 1}\mathcal{E}_{S_i}(\hat{h})\\
            =& \sum_{i \neq j}\frac{1}{M -1}|(\int_X (\mathbb{P}_i(x) - \mathbb{P}_j(x))* \ell (\hat{h}(x), y)\mathrm{d}x)| \\
            & + \sum_{i \neq j}\frac{1}{M - 1}\mathcal{E}_{S_i}(\hat{h})\\
            \leq& \sum_{i \neq j}\frac{1}{M - 1}\sup_{g \in {\rm G}}|\int_X g(x)(\mathbb{P}_i(x) - \mathbb{P}_j(x))\mathrm{d}x| \\
            & + \sum_{i \neq j}\frac{1}{M - 1}\mathcal{E}_{S_i}(\hat{h})\\
            =& \sum_{i \neq j}\frac{1}{M -1}{\rm IPM_G}(\mathbb{P}_i, \mathbb{P}_j)+ \sum_{i \neq j}\frac{1}{M - 1}\mathcal{E}_{S_i}(\hat{h}),\\
        \end{aligned}
    \end{equation}
combined with Lemma~\ref{VC}, we find that for any $\delta \in (0, 1)$, the following holds with probability at least $1- \delta$,
\begin{equation}
    \begin{aligned}
       & \mathcal{E}_{S_j}(\hat{h}) \\
       \leq& \sum_{i \neq j}\frac{1}{M - 1}\mathcal{E}_{S_i}(\hat{h}) + \sum_{i \neq j}\frac{1}{M -1}{\rm IPM_G}(\mathbb{P}_i, \mathbb{P}_j) \\
        \leq & \sum_{i \neq j}\frac{1}{M - 1}\hat{\mathcal{E}}_{S_i}(\hat{h}) + \sqrt{\frac{8d\ln{\frac{2e(M-1)n}{d}} + 8\ln{\frac{4}{\delta}}}{(M-1)n}}\\
        &+ \sum_{i \neq j}\frac{1}{M -1}{\rm IPM_G}(\mathbb{P}_i, \mathbb{P}_j)\\
        \leq & \sum_{i \neq j}\frac{1}{M - 1}\hat{\mathcal{E}}_{S_i}(h_{S_j}^*) + \sqrt{\frac{8d\ln{\frac{2e(M-1)n}{d}} + 8\ln{\frac{4}{\delta}}}{(M-1)n}}\\
        &+ \sum_{i \neq j}\frac{1}{M -1}{\rm IPM_G}(\mathbb{P}_i, \mathbb{P}_j)\\
        \leq & \sum_{i \neq j}\frac{1}{M - 1}\mathcal{E}_{S_i}(h_{S_j}^*) + 2\sqrt{\frac{8d\ln{\frac{2e(M-1)n}{d}} + 8\ln{\frac{4}{\delta}}}{(M-1)n}}\\
        &+ \sum_{i \neq j}\frac{1}{M -1}{\rm IPM_G}(\mathbb{P}_i, \mathbb{P}_j)\\
        \leq & \mathcal{E}_{S_j}(h_{S_j}^*) + 2\sqrt{\frac{8d\ln{\frac{2e(M-1)n}{d}} + 8\ln{\frac{4}{\delta}}}{(M-1)n}}\\
        &+ \sum_{i \neq j}\frac{2}{M -1}{\rm IPM_G}(\mathbb{P}_i, \mathbb{P}_j).\\
    \end{aligned}
\end{equation}

\end{proof}

\vfill

\end{document}